\documentclass{article}

\usepackage{microtype}
\usepackage{graphicx}
\usepackage{subfigure}
\usepackage{booktabs} %

\usepackage{hyperref}

\usepackage[accepted]{icml2025}
\gdef\icmlarxiv

\usepackage{amsmath}
\usepackage{amssymb}
\usepackage{mathtools}
\usepackage{amsthm}

\usepackage[capitalize,noabbrev]{cleveref}

\theoremstyle{plain}

\theoremstyle{definition}

\theoremstyle{remark}

\usepackage{multirow}
\usepackage{pifont}
\usepackage{array}
\usepackage{tikz}
\usepackage{pgfplots,pgfplotstable}
\usepackage{tabu}
\usepackage{fancyvrb}
\usepackage{listings}
\usepackage{fontawesome}
\usepackage{tablefootnote}
\usepackage{colortbl}
\usepackage{soul}
\usepackage{tcolorbox}
\usepackage{xspace}
\usepackage{caption}
\tcbuselibrary{breakable}

\usetikzlibrary{positioning}
\pgfplotsset{width=7cm,compat=1.3}

\newcommand{\OURS}{\textsc{MILS}\xspace}
\newcommand{\generator}{\textsc{Generator}\xspace}
\newcommand{\scorer}{\textsc{Scorer}\xspace}
\newcommand{\llamaOnly}{Llama\xspace}
\newcommand{\llamaNumOnly}[1]{Llama #1\xspace}
\newcommand{\llama}[2]{Llama #1 #2\xspace}

\definecolor{niceblue}{RGB}{37,165,203}
\definecolor{mustard}{RGB}{253,177,26}
\definecolor{darkred}{RGB}{175,32,67}
\definecolor{deepblue}{RGB}{0,69,122}
\definecolor{corn}{rgb}{0.98, 0.93, 0.36}
\definecolor{cosmiclatte}{rgb}{1.0, 0.97, 0.91}
\definecolor{babyblueeyes}{rgb}{0.63, 0.79, 0.95}

\newcommand{\hide}[1]{}

\newcolumntype{H}{>{\setbox0=\hbox\bgroup}c<{\egroup}@{}}
\newcommand{\PreserveBackslash}[1]{\let\temp=\\#1\let\\=\temp}
\newcolumntype{C}[1]{>{\PreserveBackslash\centering}p{#1}}
\newcolumntype{L}[1]{>{\PreserveBackslash\raggedright}p{#1}}

\makeatletter
\DeclareRobustCommand\onedot{\futurelet\@let@token\@onedot}
\def\@onedot{\ifx\@let@token.\else.\null\fi\xspace}

\def\eg{\emph{e.g}\onedot} 
\def\ie{\emph{i.e}\onedot} 
\def\cf{\emph{cf}\onedot} 
 \def\vs{\emph{vs}\onedot}

\def\etal{\emph{et al}\onedot}
\makeatother

\icmltitlerunning{LLMs can see and hear without any training}

\begin{document}

\twocolumn[
\icmltitle{LLMs can see and hear without any training}

\icmlsetsymbol{equal}{*}

\begin{icmlauthorlist}
\icmlauthor{Kumar Ashutosh}{zzz,xxx}
\icmlauthor{Yossi Gandelsman}{zzz,yyy}
\icmlauthor{Xinlei Chen}{zzz}
\icmlauthor{Ishan Misra}{zzz}
\icmlauthor{Rohit Girdhar}{zzz}
\end{icmlauthorlist}

\icmlaffiliation{zzz}{Meta AI}
\icmlaffiliation{xxx}{UT Austin}
\icmlaffiliation{yyy}{UC Berkeley}

\icmlteaser{}

\icmlcorrespondingauthor{Kumar Ashutosh}{kumar.ashutosh@utexas.edu}
\icmlcorrespondingauthor{Rohit Girdhar}{rgirdhar@meta.com}

\icmlkeywords{large language models, LLMs, multimodal, reasoning, captioning, image generation, test-time optimization, gradient-free optimization}

\vskip 0.2in
]

\printAffiliationsAndNotice{}  %

\begin{abstract}
We present \OURS: Multimodal Iterative LLM Solver, a surprisingly simple,
training-free
approach, to imbue multimodal capabilities into
your favorite LLM. %
Leveraging their innate ability
to perform %
multi-step reasoning,
\OURS prompts the LLM to generate candidate outputs, %
each of which are scored and fed back iteratively, %
eventually generating a solution to the task.
This enables various applications that typically require training specialized models on task-specific data.
In particular, we establish a new state-of-the-art on emergent zero-shot
image, video and audio captioning.
\OURS seamlessly applies %
to media generation as well, %
discovering prompt rewrites to %
improve text-to-image generation, and even edit prompts for style transfer!
Finally, being a gradient-free optimization approach, \OURS can invert multimodal embeddings into text, enabling applications like %
cross-modal arithmetic.
\end{abstract}

\section{Introduction}
\label{sec:intro}

Test-time reasoning ability of Large Language Models (LLMs) has emerged as
a powerful tool for solving challenging tasks. Recently, OpenAI introduced O1~\cite{openai2024o1}, a model trained using reinforcement learning to leverage test-time compute %
for progressively better results, especially on complex math and coding tasks. %
Even without additional training, LLMs have shown impressive improvements by using test time compute in the form Chain-of-Thought (CoT) reasoning, by rolling out an execution plan to respond to a user's query~\cite{wei2022chain,kojima2022large}.

In this work, we leverage this innate reasoning ability in LLMs to solve multimodal understanding and generation tasks, without needing any training! Our approach, \OURS: a {\bf M}ultimodal {\bf I}terative {\bf L}LM {\bf S}olver, uses LLMs as a ``{\em \generator}''
to propose candidate solutions to a given task, and an off-the-shelf multimodal
model as a ``{\em \scorer}'' to evaluate the quality of each proposal for the said task.
The output of the \scorer is fed back into the \generator to give it feedback,
and helps produce the next set of candidates that are more likely to solve the task. This iterative process is run until convergence, or a certain number of steps, and produces an output for the task. We find this simple approach is
surprisingly powerful and versatile, working across a variety of tasks and modalities.
Using different combinations of \generator{}s and \scorer{}s, \OURS
is able to tackle tasks including multimodal captioning, generation, editing, and multimodal arithmetic.

Most prior work for such tasks
uses specialized models, often trained on data curated
for that task. For instance, zero-shot image captioning models are often
still trained on paired image-caption data. \OURS, on the other hand, does not need any such training, and exhibits {\em emergent} zero-shot capabilities. For instance, for image captioning, \OURS uses a standard \llamaOnly{}~\cite{dubey2024llama} LLM as the \generator, along with CLIP~\cite{radford2021learning} as the \scorer. Note that while CLIP is trained on image-text data, it is
not trained on clean image-caption data that typical captioning models are trained on.
Most vision language models that produce captions leverage CLIP only as an initialization, and require post-training on curated captioning data. Hence, while such models may exhibit zero-shot generalization to a {\em new data distribution} at test time, \OURS exhibits {\em emergent} zero-shot generalization to the {\em new task} of captioning.

Furthermore, while there do exist some captioning approaches that do not leverage captioning data~\cite{salewski2023zero,tewel2022zerocap,shaharabany2023zero,zeng2024meacap}, they are limited to
a specific modality and more importantly, the specific task. These approaches typically leverage
gradients from CLIP to guide the next token prediction, limiting them to
captioning. \OURS, on the other hand,
seamlessly generalizes
to new tasks and modalities by simply swapping out the \generator and \scorer modules.
For instance, a \generator constructed by simply chaining an LLM to a Text-to-Image (T2I) model,
is able to improve over state-of-the-art T2I models by using the LLM as a ``prompt rewriter'',
a capability not afforded by prior work.

In this work, we show applicability of \OURS across three different visual and non-visual modalities: images, videos and audio, and three different tasks: captioning, generation and editing. Additionally, we show that \OURS, being a gradient-free approach, can be used to invert %
multimodal embeddings into discrete text. This is in contrast to prior work~\cite{kazemi2024we} that uses gradient-based inversion of embeddings into continuous spaces like images. This ability enables
novel applications, \eg multimodal arithmetic, by inverting multimodal samples into text, combining them, and mapping them back
using \OURS's generation ability. We visualize some of these capabilities in~\cref{fig:teaser}.

\section{Related Work}
\label{sec:relwork}

{\noindent \bf Multimodal embedding} spaces are typically learned by collecting
large amounts of paired multimodal data from the internet, often images and text,
and learning encoders for each modality using a pairwise similarity objective~\cite{radford2021learning,ilharc2021openclip,zhai2023sigmoid,li2023blip}.
Such models can further
be expanded to additional modalities by collecting text-paired data~\cite{wang2023internvid,guzhov2022audioclip}, or data paired to a different modality
in that embedding space~\cite{girdhar2023imagebind,gong2023mmg}. These embeddings
have enabled various applications, including zero-shot recognition~\cite{radford2021learning,girdhar2023imagebind} open-world object detection~\cite{zhou2022detecting}, and even image generation~\cite{ramesh2022hierarchical}.
We leverage them to compute a similarity score across modalities,
which helps guide the optimization, imbuing multimodal capabilities into an otherwise blind and deaf LLM.

{\noindent \bf Generative models}
have recently gained popularity due to their ability to generalize
to new tasks, often zero-shot. LLMs~\cite{dubey2024llama,jiang2023mistral,team2024gemma} have emerged as the model of choice for discrete input, such as text. Owing to their large size and training on massive corpuses,
followed by instruction tuning on high quality data often involving human feedback,
such models are powerful tools for a variety of tasks.
Approaches such as chain-of-thought prompting~\cite{wei2022chain,kojima2022large,menon2024whiteboard}, and more recently,
training the LLM for better reasoning~\cite{openai2024o1}, have further improved their performance on complex math and coding tasks. %
However, instruction tuning involves training or finetuning LLMs for the target task and modality. In contrast, \OURS is an inference-time method that does not require any training and optimizes for the solution at runtime.
Recent work has even leveraged LLM's reasoning ability iteratively, to solve optimization~\cite{yang2023large} and generation~\cite{manas2024improving} tasks. However, they are either not evaluated on visual tasks~\cite{yang2023large}, or shown to have \emph{emergent} zero-shot capabilities~\cite{manas2024improving}.
Another class of generative models involve diffusion~\cite{nichol2021improved} or flow-matching~\cite{lipman2022flow}, popular for continuous domains such as images~\cite{rombach2022high,betker2023dalle3,dai2023emu,saharia2022photorealistic} and videos~\cite{polyak2024movie,girdhar2024emu,ho2022imagen,blattmann2023align}.
Such models have dramatically improved in media generation capability, often leveraging LLMs too, for better training data captioning, and inference-time prompt rewrites~\cite{betker2023dalle3,polyak2024movie}.

{\noindent \bf Zero-shot multimodal understanding} is studied in two forms: zero-shot
across data distributions, and {\em emergent} zero-shot~\cite{girdhar2023imagebind}, where a model generalizes to completely new tasks and not just new data.
Multimodal variants of popular LLMs~\cite{dubey2024llama,agrawal2024pixtral,li2023m} fall in the former bucket, as they are typically trained or tuned on the kind of data seen at test time. Our focus in this work is the latter, as we show \OURS generalizes
to completely new tasks at test time. Prior work~\cite{tewel2022zerocap,conzic,zeng2024meacap,salewski2023zero,shaharabany2023zero} has attempted this setting, however for specific modalities using specialized techniques. \OURS, on the other hand, %
seamlessly generalizes to many different modalities, across understanding and generation tasks. %

\section{\OURS}

\begin{figure}[t]
    \centering
    \includegraphics[width=\linewidth]{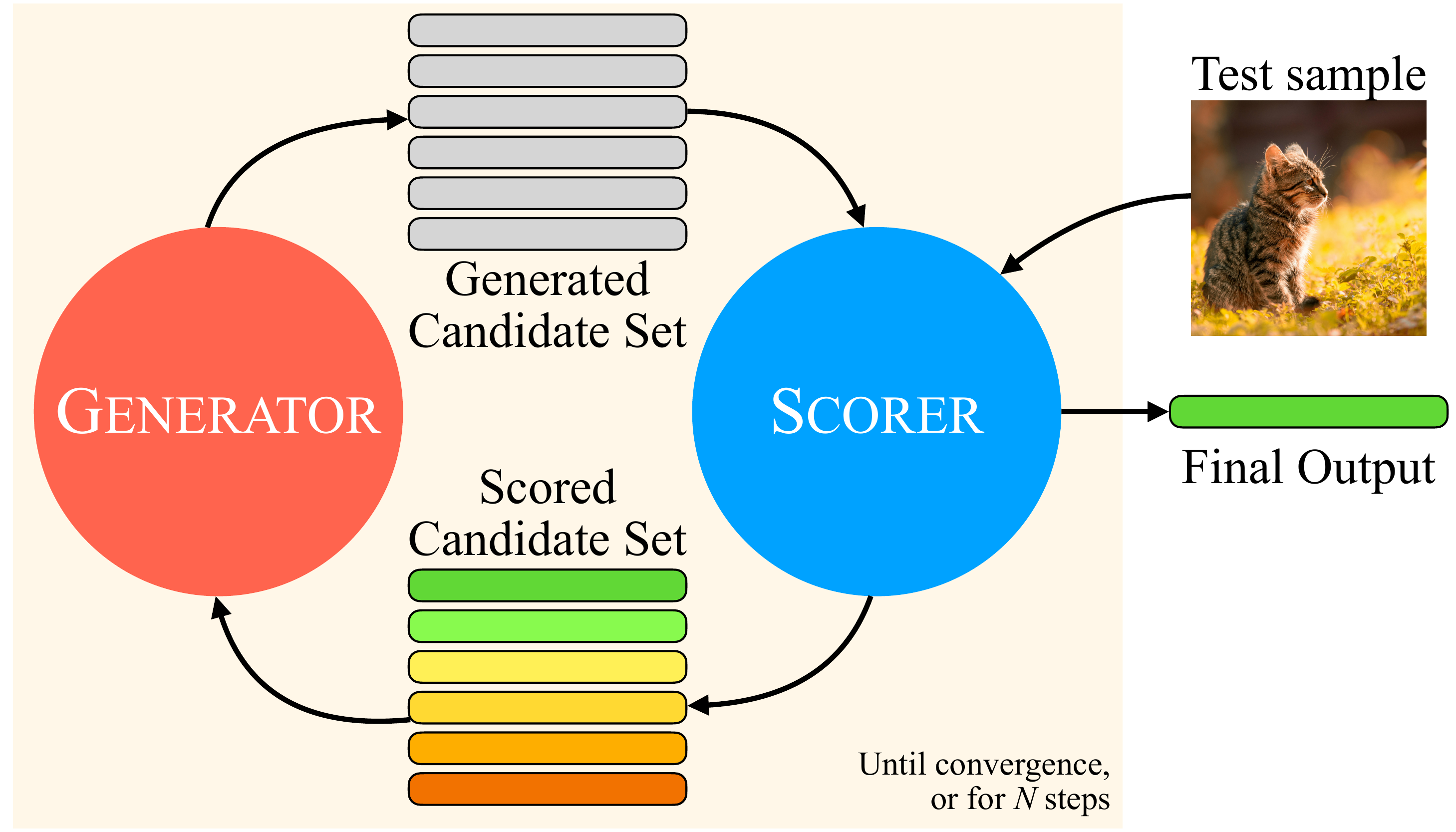}
    \caption{
        {\bf \OURS} leverages two key modules, \generator and \scorer, to solve multimodal tasks. %
        The \generator will generate a number of
        text candidates, \eg captions for image captioning and prompts for T2I, each of which will be scored by the \scorer, and passed back into the \generator as feedback to generate the next batch of text candidates, eventually producing the final output for the input test sample.
    }
    \label{fig:approach}
\end{figure}

We now describe our simple approach for solving multimodal tasks using \OURS.
Since it is training-free, \OURS only takes the test sample as input.
It relies on two key modules, referred to as the \generator and the \scorer. As the names suggest, \generator generates candidate solutions for the task, while the \scorer scores each of those candidates, and sends them back to the \generator to generate an improved candidate set.
For certain tasks, this process may be bootstrapped with scores on an initial candidate set. %
This optimization is run until convergence or a fixed number of iterations, and produces the final solution to the task. \cref{fig:approach} illustrates the overall approach.

{\noindent \bf \generator.}
The goal of the \generator is to produce candidate outputs $\mathcal{C}$, that solve a given task.
It takes as input some text, $\mathcal{T}$, which contains a description of the task, along with
scores $\mathcal{S}$ (if any) from the \scorer for the previous optimization step.
It leverages this
signal to produce the next set of candidate generations.
The \generator is typically modeled using an LLM, given its ability to take text as input and reason over it. The output, however, is not limited to text. The candidate generations can be used to prompt a subsequent model to generate other modalities, such as using a text-to-image (T2I)
model like Emu~\cite{dai2023emu}
to generate images. Some \generator{}s could also use the test sample as input, \eg for tasks like image editing or stylization.

{\noindent \bf \scorer.}
The goal of the \scorer is to compute a scalar score $\mathcal{S}\!\in\!\mathbb{R}$, for the candidates $\mathcal{C}$ from the \generator. It takes as input the test sample, along with $\mathcal{C}$, and compares them. A \scorer{} could be implemented in various different ways. For instance, it could be a low-level image processing function comparing textures in two images, or it could be a learned machine learning model, such as CLIP~\cite{radford2021learning,ilharc2021openclip}. The \scorer sorts all the
candidates based on their score, and returns the top-$K$ candidates along with scores. Depending on the \generator{}'s capacity (context length), the scorer may return the full list of scores, or use an $\epsilon$-greedy strategy to include some low-scoring candidates. In initial experiments, we found greedy top-$K$ to perform the best, hence use that in this work. The output is formatted into the text $\mathcal{T}$, and passed back to the \generator.

{\noindent \bf Optimization process.}
\OURS searches for the optimal generation $\mathcal{C}$ under the
\scorer{}'s cost function. The optimization process is run for $N$
steps, or until convergence. Convergence can be defined by the
similarity of the candidate set $\mathcal{C}$ over successive steps.
Depending on the task, the optimization process can be bootstrapped by an initial candidate set of generations and scoring them. For instance, in case of image captioning, it could simply be a large set of possible image captions from the \generator. For other tasks like T2I, one does not need such an initial set.

\section{Experiments}\label{sec:expts}

\begin{table}[t]
    \setlength{\tabcolsep}{3pt}
    \renewcommand{\arraystretch}{1.2}
    \centering
    \resizebox{\linewidth}{!}{
    \begin{tabular}{|c|C{1.5in}|C{1.5in}|C{1.5in}|} %
        \hline
        & \frame{\includegraphics[width=1.5in, trim={0 5cm 0 5cm}, clip]{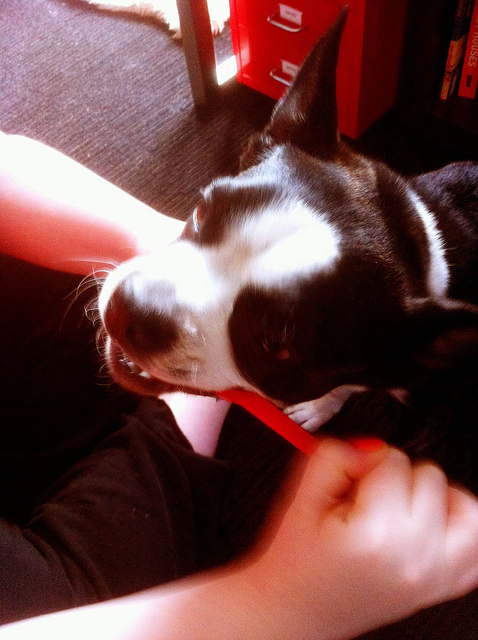}}
        & \frame{\includegraphics[width=1.5in, trim={0 2cm 0 5.7cm}, clip]{figs/main_results/image_cap/COCO_val2014_000000032811.jpg}}
        & \frame{\includegraphics[width=1.5in]{figs/main_results/image_cap/COCO_val2014_000000131138.jpg}}
        \\
        \hline
        \rowcolor{cosmiclatte}
        \rotatebox[origin=r]{90}{\scriptsize{\OURS}}
        & {\it Boston Terrier's sharp teeth firmly gripping a rope toy tightly.}
        & {\it Close-up of a bird near a ship's cabin window.}
        & {\it Cluttered desk with a large monitor and a window.}
        \\
        \hline
        \rotatebox[origin=r]{90}{\scriptsize{MeaCap~\cite{zeng2024meacap}}}
        & {\it The image depicts that dog licking a man who is in inappropriately close proximity.}
        & {\it The image depicts that observation boat in the water below us, " he said.}
        & {\it The image depicts that workspace during an Cluttered desk with a large monitor and a online career as a consultant.}
        \\
        \hline
    \end{tabular}%
    }
    \captionof{figure}{
        \textbf{Image Captioning using \OURS}, compared to existing state-of-the-art zero-shot
        approach, MeaCap~\cite{zeng2024meacap}. \OURS, while being a much simpler approach, produces more accurate and syntactically correct captions to the image.
    }
    \label{fig:results:image_caps}
\end{table}

We now empirically evaluate \OURS and compare it to
existing approaches on some of the multimodal understanding
and generation tasks enabled by it.
For each of the downstream applications, we describe the \generator, \scorer, benchmarks and evaluation setup, followed by the key results.
Finally in~\cref{sec:expts:ablations} we ablate the various design choices in \OURS.

Note that \OURS is a test-time optimization method and exhibits
{\em emergent zero-shot} behavior~\cite{girdhar2023imagebind}, generalizing not only to a new test data distribution, but to the new task and modality itself. This is contrast to most existing zero-shot work that typically needs
task/modality-specific data curation or training.
Since most prior work is of the latter type, it is hard to perform fair comparisons.
Nevertheless, we compare to
the closest zero-shot approaches and show that \OURS is competitive or better, even compared to
methods tuned for that specific task or modality.

\begin{table}[t]
    \centering
    \resizebox{\linewidth}{!}{
        \begin{tabu}{l|HHccccHHHHHH}
            \toprule[1pt]
            Method &Training & Memory &BLEU$_{4}$ & CIDEr & METEOR & SPICE &CLIP-S &BLIP2-S & In &Near &Out &Overall  \\
            \midrule
            ZeroCap~\cite{tewel2022zerocap} &\ding{55} & \ding{55}&2.6  &14.6 &11.5 &5.5 &\underline{0.87} &0.70 &13.3 &14.9 &19.7 &16.6\\
            ConZIC~\cite{conzic} &\ding{55} & \ding{55}&1.3  &13.3 &11.2 &5.0 &\textbf{1.00} &\underline{0.76} &15.4 &16.0 &20.3 &17.5  \\
            CLIPRe~\cite{li2023decap} &\ding{55} & CC3M &4.6  &25.6 &13.3 &9.2&0.84 &0.70 &23.3 &26.8 &36.5 &28.2 \\
            \rowfont{\color{gray}} MeaCap$_{\mathrm{TF}}$~\cite{zeng2024meacap} &\ding{55} &CC3M & 7.1  & 42.5 &16.6 & 11.8 &0.84 &\textbf{0.81} &\underline{35.3} & \underline{39.0} &45.1 &\underline{40.2}  \\
            MeaCap$_{\mathrm{TF}}^*$~\cite{zeng2024meacap} & && 4.5 & 26.0 & 14.1 & 9.4 & ?? & ?? \\
            \midrule
            \OURS & && {\bf 8.0} & {\bf 33.3} & {\bf 15.0} & {\bf 9.6} & ?? & ??\\
            \bottomrule[1pt]
        \end{tabu}%
        }
    \caption{{\bf Zero-shot image captioning} on MSCOCO~\cite{karpathy2015deep}.
    Despite being far simpler than existing approaches, \OURS performs competitively on all automatic metrics, and especially METEOR and SPICE which take into account
    the semantic similarity. $^*$refers to results we obtained by running the provided code. %
    }
    \label{table:zero-shot-image-caption}
\end{table}

\subsection{Image Captioning}\label{sec:expts:image_caption}

We start with the fundamental image understanding task of producing a textual caption for a given image.

{\noindent \bf \generator.}
We use the \llama{3.1}{8B}~\cite{dubey2024llama} LLM as the core generation module. We generate an initial list of 30K prompts that we use to bootstrap the optimization process.
To ensure diversity in this initial set, we prompt the LLM with different object categories to generate a list of prompts, and combine them, similar to~\cite{gandelsman2024interpreting}. %
Then for each optimization step, we keep the top-50 highest scoring generations from the \scorer, and convert them into a textual prompt.
The prompt used is described in~\cref{sec:appdx:llm_prompts}.
We run the optimization process for 10 steps.

{\noindent \bf \scorer.}
We score the candidate captions against the test image using an image-text similarity model, %
SigLIP~\cite{zhai2023sigmoid}. %
Note that unlike image captioning models that leverage curated image-text pairs~\cite{karpathy2015deep},
SigLIP, %
by itself is not capable of captioning~\cite{shen2021much}.
Nevertheless, combined with \OURS, it can serve as an effective captioner, as shown next.

\begin{table}[t]
    \centering
    \resizebox{\linewidth}{!}{
        \begin{tabu}{l|c|Hcc}
            \toprule[1pt]
            Method & Training Data & BLEU$_{4}$ & CIDEr & METEOR  \\
            \midrule
            \rowfont{\color{gray}} Nagrani \etal~\cite{nagrani2022learning} & HowTo100M~\cite{miech19howto100m} & 7.5 & 0.5 & 8.23\\
            \rowfont{\color{gray}} Nagrani \etal~\cite{nagrani2022learning} & VideoCC3M~\cite{nagrani2022learning} & {\bf 13.2} & {\bf 8.2} & 11.3 \\
            \midrule
            \OURS & -- & 3.0 & 2.3 & {\bf 14.4} \\
            \bottomrule[1pt]
        \end{tabu}%
        }
    \caption{{\bf Zero-shot video captioning} on MSR-VTT~\cite{xu2016msr}.
        \OURS outperforms~\cite{nagrani2022learning} when trained on HowTo100M, and is competitive to it when trained on the much cleaner VideoCC3M dataset, outperforming it on METEOR.
        We {\color{gray} grayed}~\cite{nagrani2022learning} since it is
        trained for video captioning, while \OURS is not.
    }
    \label{table:zero-shot-video-caption}
\end{table}

{\noindent \bf Benchmarks and Metrics.}
We evaluate \OURS on the MSCOCO captioning test set~\cite{karpathy2015deep}. It consists of 5,000 images sampled from the MSCOCO dataset~\cite{lin2014microsoft}. We use the standard suite of captioning evaluation metrics, including BLEU~\cite{papineni2002bleu}, METEOR~\cite{banerjee2005meteor},
CIDEr~\cite{vedantam2015cider} and SPICE~\cite{anderson2016spice}.
We focus our attention on the
METEOR and SPICE metrics, as those take into account semantic similarity rather than exact word match, and are better correlated with human preference~\cite{anderson2016spice}.
This is particularly important for emergent zero-shot approaches like \OURS, which are not trained
to learn the vocabulary used in a given benchmark or modality.

{\noindent \bf Results.}
We compare \OURS to existing baselines in~\cref{table:zero-shot-image-caption}. Some of the baselines, such
as ZeroCap~\cite{tewel2022zerocap}, also leverage language models in conjunction with a CLIP-like model. However, they propose a gradient based
optimization process to search for the optimal next token given a current generation.
Other approaches like MeaCap~\cite{zeng2024meacap} filter key concepts from a memory module, and leverage a number of text and
multimodal encoders in a multi-step process to produce the caption.
In contrast, \OURS is simpler conceptually and to implement, while obtaining better results.
We also show examples of the captions generated using \OURS, and compare them to MeaCap in~\cref{fig:results:image_caps}. %
\OURS, without ever having seen any captioning data or having captioning-specific training %
is able to generate faithful and syntactically correct captions. %

\begin{table}[t]
    \centering
    \resizebox{\linewidth}{!}{
        \begin{tabular}{l|HHHHHccHcc}
            \toprule[1pt]
            Method & BLEU$_{4}$ & ROUGE$_{\mathrm{L}}$ & CIDEr & METEOR & SPICE & BLEU$_{4}$ & ROUGE$_{\mathrm{L}}$ & CIDEr & METEOR & SPICE  \\
            \midrule
            ZerAuCap~\cite{salewski2023zero} & 6.8 & 33.1 & 28.1 & 12.3 & 8.6 & {\bf 2.9} & {\bf 25.4} & 14.0 & 9.4 & 5.3 \\
            \OURS & & & & & & 2.7 & 23.1 & 8.9 & {\bf 12.4} & {\bf 7.6} \\
            \bottomrule[1pt]
        \end{tabular}%
        }
    \caption{{\bf Zero-shot audio captioning} on Clotho~\cite{drossos2020clotho} dataset.
    \OURS performs competitively to the existing zero-shot audio captioning approach ZerAuCap, %
    even outperforming it on semantics-aware metrics like METEOR and SPICE,
    while being simpler and applicable to many other modalities and tasks.
    }
    \label{table:zero-shot-audio-caption}
\end{table}

\subsection{Video Captioning}\label{sec:expts:video_caption}

Owing to its simplicity
and versatility, \OURS seamlessly transfers to videos without
major changes. We use the same \generator as described for image captioning in~\cref{sec:expts:image_caption}, along with the same initial prompt set. For \scorer, we use a ViCLIP~\cite{wang2023internvid} ViT-L/14 model that operates on 8 frames from the video, and returns a similarity score between the video and the caption. We experiment on the MSR-VTT~\cite{xu2016msr} test set, which contains 2,990 videos, each between 10 to 30 seconds long. We report our results in~\cref{table:zero-shot-video-caption}. Since most prior work in video captioning leverages video-caption training data, we compare \OURS to~\cite{nagrani2022learning}, which learns a vision-language model on the HowTo100M~\cite{miech19howto100m} or the VideoCC3M~\cite{nagrani2022learning} datasets, and reports performance on
MSR-VTT. We use the CIDEr~\cite{vedantam2015cider} and METEOR~\cite{banerjee2005meteor} metrics as reported in the prior work. We find that \OURS, in spite of never having been trained for video captioning,  outperforms~\cite{nagrani2022learning} trained on HowTo100M on both metrics. Even when compared to the same model trained on the much cleaner VideoCC3M data, \OURS outperforms it on the semantics-aware METEOR metric. %
This difference in performance of the baseline in the two settings shows the importance of training data for video captioning models. \OURS, being competitive with these without needing any video captioning training, is very promising. We show qualitative results in the~\cref{sec:appdx:addl_qual}.

\begin{figure}
    \centering
    \includegraphics[width=\linewidth]{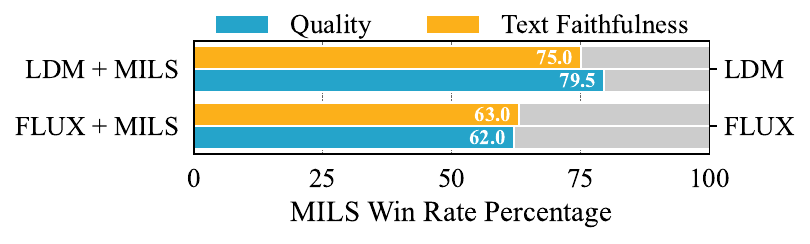}
    \caption{
        {\bf Improved text-to-image (T2I) generation using \OURS.}
        We apply \OURS to two of the latest, state-of-the-art T2I models, a latent diffusion model (LDM), and FLUX.1 [schnell] (FLUX).
        We compare \OURS{}'s outputs to the generations from the initial models using human annotators. Evaluated
        over the 200 prompt DrawBench %
        dataset, the annotators clearly preferred \OURS{}'s generations on both overall quality and text faithfulness, across
        both models.
    }\label{fig:results:imagegen_graph}
\end{figure}

\subsection{Audio Captioning}\label{sec:expts:audio_caption}

Similar to videos, \OURS transfers seamlessly to audio captioning as well. We use the same \generator as in~\cref{sec:expts:image_caption}, along with 50K initial audio prompts generated using an LLM (\cf~\cref{sec:appdx:llm_prompts}).
As the \scorer, we use the ImageBind~\cite{girdhar2023imagebind} model, that maps multiple modalities, including audio and text, to a shared embedding space. %
We evaluate our approach on a popular audio captioning dataset, Clotho~\cite{drossos2020clotho}. %
We use automatic captioning metrics as used in prior work, and described in~\cref{sec:expts:image_caption}.
We report our performance in~\cref{table:zero-shot-audio-caption}. \OURS obtains strong performance \vs a comparable zero-shot approach, ZerAuCaps~\cite{salewski2023zero}, outperforming it
especially on semantics-aware metrics like METEOR and SPICE.
While other approaches for audio captioning have been proposed, they require training on
audio-caption data~\cite{kong2024audio}.
See~\cref{sec:appdx:addl_qual} for qualitative results.

\begin{table}[t]
    \setlength{\tabcolsep}{2pt}
    \centering
    \resizebox{\linewidth}{!}{
    \begin{tabular}{|C{0.1in}|C{1.5in}|C{1.5in}|C{1.5in}|} %
        \hline
        \rotatebox[origin=r]{90}{\scriptsize{Prompt}}
        & {\it \small A triangular purple flower pot. A purple flower pot in the shape of a triangle.}
        & {\it \small Five cars on the street.}
        & {\it \small Appliance or compartment which is artificially kept cool \& used to store food \& drink.}
        \\
        \hline
        \rotatebox[origin=c]{90}{\scriptsize{LDM}}
        & \raisebox{-0.5\height}{\frame{\includegraphics[width=1.5in]{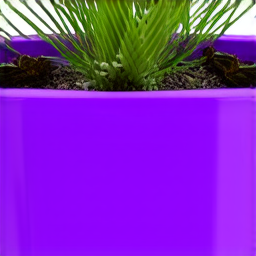}}}
        & \raisebox{-0.5\height}{\frame{\includegraphics[width=1.5in]{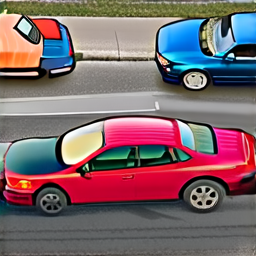}}}
        & \raisebox{-0.5\height}{\frame{\includegraphics[width=1.5in]{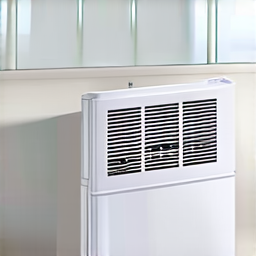}}}
        \\[1.9cm]
        \rowcolor{cosmiclatte} \rotatebox[origin=c]{90}{\scriptsize{LDM + \OURS}}
        & \raisebox{-0.5\height}{\frame{\includegraphics[width=1.5in]{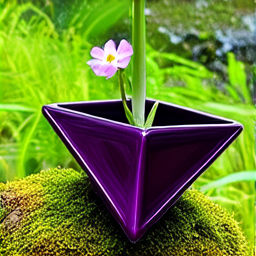}}}
        & \raisebox{-0.5\height}{\frame{\includegraphics[width=1.5in]{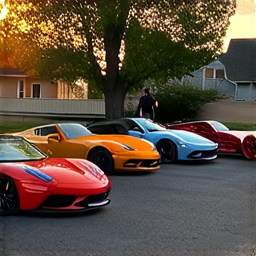}}}
        & \raisebox{-0.5\height}{\frame{\includegraphics[width=1.5in]{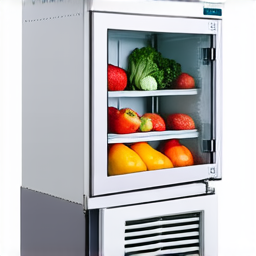}}}
        \\
        \hline
    \end{tabular}%
    }
    \captionof{figure}{
        \textbf{Improving image generation using \OURS.}
        Applying \OURS to a \generator using the same base model,
        a Latent Diffusion Model (LDM)
        in this case, leads to much higher quality images. We show the original input
        prompt, the generation from the base model, and from \OURS.
    }
    \label{fig:results:imagegen}
\end{table}

\subsection{High-Quality Image Generation}\label{sec:expts:image_gen}

As mentioned earlier, \OURS is not limited to multimodal understanding tasks discussed so far.
We now describe how \OURS can be used for multimodal generation tasks, starting with improving text-to-image (T2I) generation models. %

{\noindent \bf \generator.}
To generate high quality images, we chain an LLM to a T2I model. Specifically, we experiment with two state-of-the-art models,
a Latent Diffusion Model (LDM)~\cite{rombach2022high}
and FLUX.1 [schnell]~\cite{flux1schnell}.
The goal of the LLM  is to ``rewrite'' the prompt passed into the T2I model, such that the final generated image improves in quality, while maintaining or improving the faithfulness to the original text prompt. Note that this \generator does not need an initial prompt set to bootstrap from.

{\noindent \bf \scorer.}
We score the generations using PickScore~\cite{kirstain2023pick}. PickScore
is a CLIP-style model takes as input an image and a text prompt, and predicts the likelihood of that image to be preferred by humans for that prompt.
We score each of the \generator outputs using PickScore along with the input prompt,
and return the scores for each of the generations. Rest of the
process proceeds the same as before. %
\begin{table}[t!]
    \setlength{\tabcolsep}{2pt}
    \centering
    \resizebox{\linewidth}{!}{
    \Large
    \begin{tabular}{cc@{\hskip 0.1in}|@{\hskip 0.1in}c}
        Test sample & Style Image & Output \\
        \frame{\includegraphics[height=1.5in]{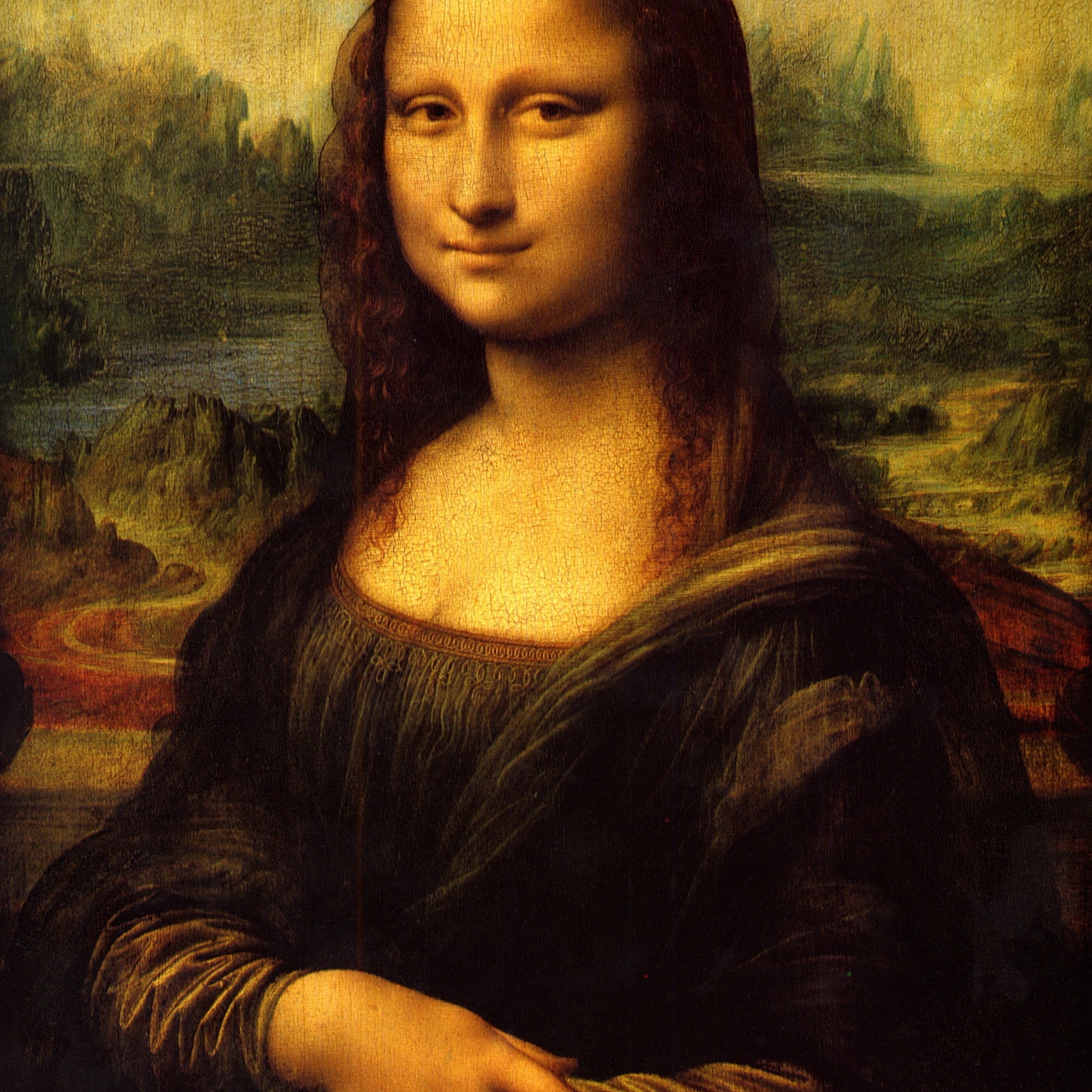}} &
        \frame{\includegraphics[height=1.5in]{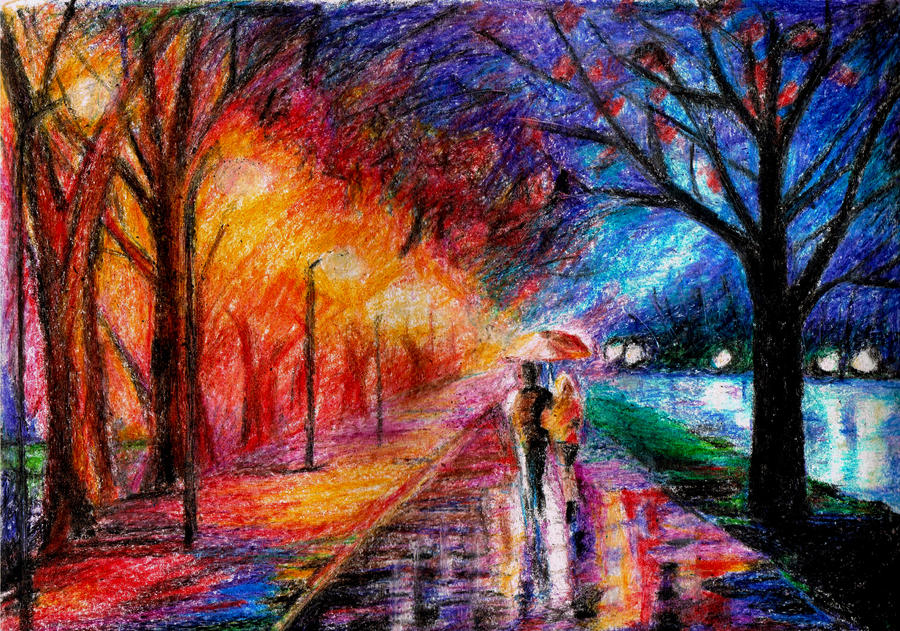}} &
        \frame{\includegraphics[height=1.5in]{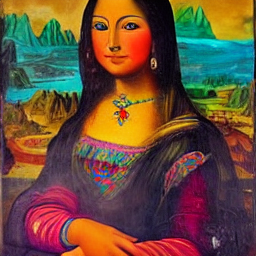}}
        \\
        \frame{\includegraphics[height=1.5in]{figs/main_results/style_transfer/monalisa-cc.jpg}} &
        \frame{\includegraphics[height=1.5in, trim={2cm 0 2cm 0}, clip]{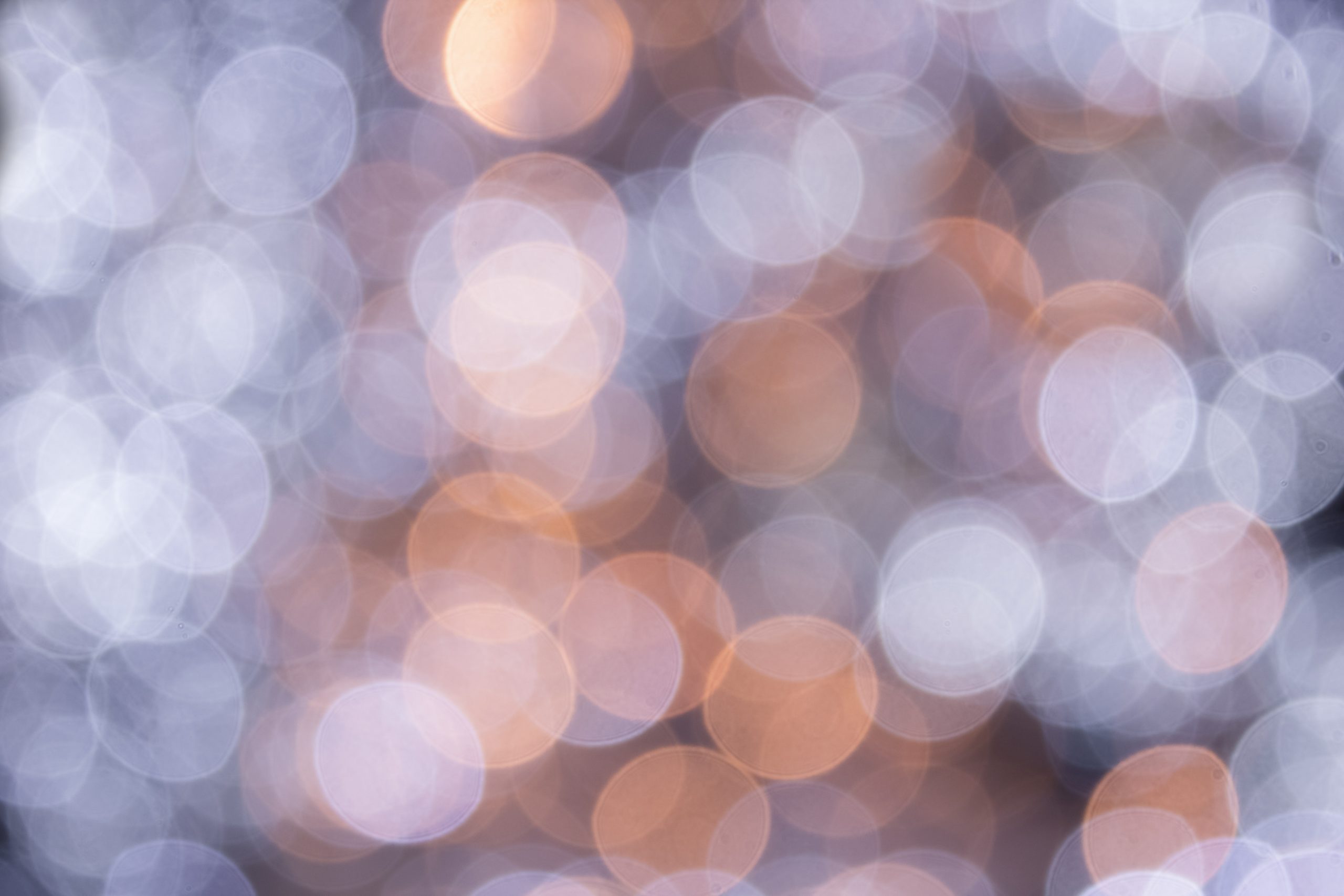}} &
        \frame{\includegraphics[height=1.5in]{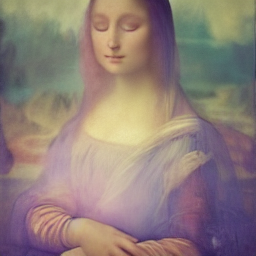}}
        \\
    \end{tabular}%
    }
    \captionof{figure}{
        \textbf{Style Transfer.} Using Gram Matrix distance~\cite{gatys2015neural} as the \scorer, \OURS
        can discover the edit prompt required to apply a given style to
        an image.
    }
    \label{fig:results:style_tx}
\end{table}

{\noindent \bf Benchmarks and Metrics.}
We use the DrawBench prompt set from Imagen~\cite{saharia2022photorealistic}, that contains 200 textual prompts, to evaluate our generations.
Due to the noisiness of automatic metrics for media generation~\cite{girdhar2024emu,ge2024content,jayasumana2024rethinking}, we perform our evaluations using human annotators from Amazon Mechanical Turk.
We follow the JUICE framework~\cite{girdhar2024emu}. In-keeping with the standard practice in media generation~\cite{dai2023emu,girdhar2024emu}, we evaluate the generations on two axes: quality or visual appeal, and the faithfulness to the input text. For each axis, we set up a web interface where an annotator is shown two images, generated either by a baseline model, or enhanced using \OURS. When evaluating for text faithfulness, the annotator is also shown the original text prompt. The annotator has to pick which image they prefer.
Each image is annotated by three annotators and we use the majority vote over the annotators to compute a win\% for each model.
See~\cref{sec:appdx:human_eval} for full human evaluation details.

{\noindent \bf Results.}
We summarize the results of our human study in~\cref{fig:results:imagegen_graph}. As seen by the win-rates, human annotators clearly prefer generations enhanced by \OURS over the generations from the base model itself. We also show qualitative comparisons in~\cref{fig:results:imagegen}, where the improvement in aesthetic quality using \OURS is clearly apparent.
We find that \OURS is able to simplify complex prompts and add aesthetic details, that improve the overall quality and faithfulness of the generations.
Recent work has shown that LLM-based prompt rewrites~\cite{betker2023dalle3,polyak2024movie} can improve media generation performance. However, they require a laborious process of manually trying various different rewrite prompts, until one finds a rewrite that works well with that model, across all test cases. \OURS can automate and complement that process, either by prompt-engineering each generation, or proposing candidate rewrites for an expert prompt engineer to improve upon.

Note that this capability is not easily afforded by prior work we compared to on earlier tasks~\cite{tewel2022zerocap,salewski2023zero}. Those methods would require computing gradients through a
multi-step diffusion process to estimate which tokens the LLM should produce next.
\OURS, being a %
gradient-free optimization approach,
easily enables such diverse applications within a simple framework.

\begin{table}[t!]
    \LARGE
    \setlength{\tabcolsep}{2pt}
    \centering
    \resizebox{\linewidth}{!}{
    \Large
    \begin{tabular}{ccp{1in}cc}
        \raisebox{-.5\height}{\frame{\includegraphics[height=1.5in, trim={0 0 2.5cm 0}, clip]{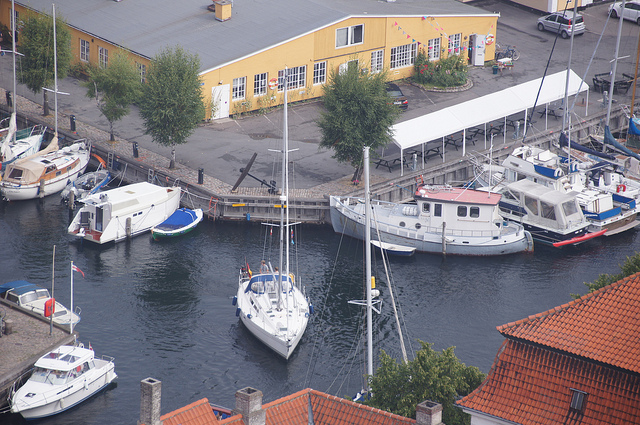}}} & + & \faVolumeUp{} Birds chirping & =
        \raisebox{-.5\height}{\frame{\includegraphics[height=1.5in]{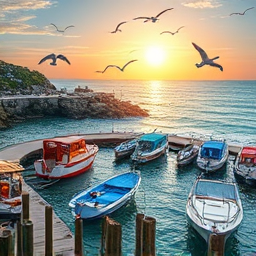}}}
        \vspace{0.2cm}
        \\
        \raisebox{-.5\height}{\frame{\includegraphics[height=1.5in]{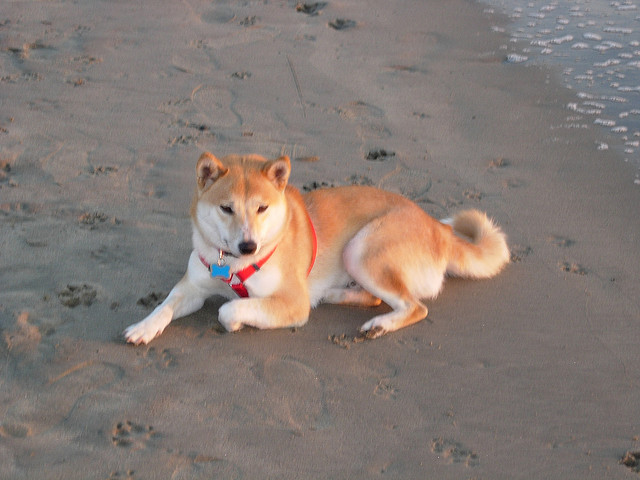}}} & + & \faVolumeUp{} People talking & =
        \raisebox{-.5\height}{\frame{\includegraphics[height=1.5in]{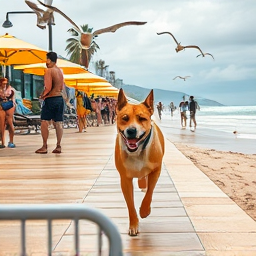}}} \\
    \end{tabular}%
    }
    \captionof{figure}{
        \textbf{\OURS enables cross-modal arithmetic} by
        inverting modalities into text, combining them, and mapping them back
        to an image.
    }
    \label{fig:results:arithmetic}
\end{table}

\subsection{Style Transfer}
Going beyond image generation, \OURS can also be applied to image editing tasks. Here we specifically consider the task of style transfer, where given a test image and a style image, the goal is to generate an image that contains the content from the test image, in the style of the style image.

{\noindent \bf \generator.}
Similar to~\cref{sec:expts:image_gen}, we implement the \generator by
chaining the output of an LLM to an image generation model.
Different from~\cref{sec:expts:image_gen}, since we want to generate the same content as in the test sample,
the \generator also takes the test sample as an input. Hence, we use
an image editing model~\cite{sheynin2024emu} as the image generation module.
It produces the stylized image given the test sample and the edit prompt from the LLM.

{\noindent \bf \scorer.}
To measure the quality of the style transfer,
we use a simple approach to estimate the similarity of colors and textures in the generated image compared to the style image. %
We use the distance between Gram matrices of the image features, as proposed in~\cite{gatys2015neural}. We compute this distance over features from different layers of a VGG19~\cite{simonyan2015very} CNN, where the lower layers ensure stylistic faithfulness, and the higher layers ensure content faithfulness.
We use \OURS to minimize both the style and content losses. %

\begin{table}[t]
    \setlength{\tabcolsep}{2pt}
    \centering
    \resizebox{\linewidth}{!}{
    \begin{tabular}{|C{0.15in}|C{1.5in}|C{1.5in}|}
        \hline
        \#
        & \includegraphics[width=1.5in, trim={0 0.85cm 0 0.85cm}, clip]{figs/ablations/image_captioning_by_step/COCO_val2014_000000198641.jpg}
        & \includegraphics[width=1.5in]{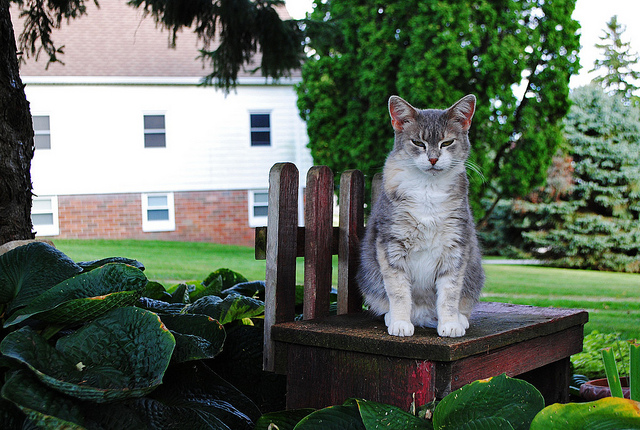}
        \\
        \hline
        1
        & {\it Photo taken with a computer in the background.}
        & {\it Picture taken in a country setting.}
        \\
        \hline
        4
        & {\it Photograph of a cat sitting on a monitor.}
        & {\it Photo of a cat in a farm setting with flowers.}
        \\
        \hline
        7
        & {\it Cat's face centered on a monitor with a nearby keyboard.}
        & {\it Photo of a cat in a farm setting with a garden bench.}
        \\
        \hline
        10
        & {\it Feline sitting on a keyboard with a nearby monitor and laptop.}
        & {\it Photo of a cat in a garden bench with a farm backdrop.}
        \\
        \hline
    \end{tabular}%
    }
    \captionof{figure}{
        \textbf{Image captioning over steps.}
        This shows the captions generated by \OURS at different steps (\#) of optimization
        process, getting progressively more accurate.
    }
    \label{fig:ablation:image_caps_by_steps}
\end{table}

\begin{figure*}[h]
    \begin{minipage}[t]{.69\textwidth}
        \Huge
        \resizebox{\linewidth}{!}{
        \begin{tikzpicture}
            \pgfplotsset{
                scale only axis,
                xmin=0, xmax=10
            }
            \begin{axis}[
                xmajorgrids,
                xminorgrids,
                grid style={line width=.1pt, draw=gray!10},
                major grid style={line width=.2pt,draw=gray!50},
                minor tick num=2,
                axis x line*=bottom,
                axis y line*=left,
                height=2in,
                width=\linewidth,
                ylabel style= {align=center},
                ylabel={\color{niceblue}{CLIP similarity}},
                xlabel={Steps},
                ytick={0.0,0.1,0.2},
            ]
            \addplot[mark=o, line width=1mm, mark options={scale=3}, smooth, tension={0.3}, niceblue] plot coordinates {
                (0, 0.01421687289538022)
                (1, 0.1272442454956472)
                (2, 0.1401589315906167)
                (3, 0.14938828766345977)
                (4, 0.1561347666159272)
                (5, 0.16111914109438658)
                (6, 0.16525025141984223)
                (7, 0.16861245894432067)
                (8, 0.1713299216926098)
                (9, 0.17369869603961707)
                (10, 0.17575965739041566)
            };\label{plot_one}
            \end{axis}

            \begin{axis}[
                axis y line*=right,
                axis x line=none,
                ylabel=y-axis 2
                grid style={line width=.1pt, draw=gray!10},
                major grid style={line width=.2pt,draw=gray!50},
                minor tick num=2,
                height=2in,
                width=\linewidth,
                ylabel style= {align=center},
                ylabel={\color{mustard}{SPICE}},
                legend style={cells={align=left}, at={(0.9,0.1)},anchor=south east},
            ]
            \addlegendimage{/pgfplots/refstyle=plot_one}\addlegendentry{CLIP Similarity}
            \addplot[mark=square, line width=1mm, mark options={scale=3}, smooth, tension={0.3}, mustard]
                coordinates{
                    (0, 5.8314)
                    (1, 8.7049)
                    (2, 9.2636)
                    (3, 9.2522)
                    (4, 9.4283)
                    (5, 9.5071)
                    (6, 9.5984)
                    (7, 9.6251)
                    (8, 9.6475)
                    (9, 9.7484)
                    (10, 9.6791)
            };
            \addlegendentry{SPICE}
            \end{axis}
        \end{tikzpicture}%
        \hfill
        \begin{tikzpicture}
            \pgfplotsset{
                scale only axis,
                xmin=0, xmax=19
            }
            \begin{axis}[
                xmajorgrids,
                xminorgrids,
                grid style={line width=.1pt, draw=gray!10},
                major grid style={line width=.2pt,draw=gray!50},
                minor tick num=2,
                axis x line*=bottom,
                axis y line*=left,
                height=2in,
                width=\linewidth,
                ylabel style= {align=center},
                ylabel={\color{niceblue}{PickScore}},
                xlabel={Steps},
            ]
            \addplot[mark=o, line width=1mm, mark options={scale=3}, smooth, tension={0.5}, niceblue] plot coordinates {
                (0, 0.21231003426015377)
                (1, 0.2158956471032488)
                (2, 0.21937074868381035)
                (3, 0.22114125204458834)
                (4, 0.2223209275938569)
                (5, 0.22323952861502755)
                (6, 0.22401689907908431)
                (7, 0.22461664003878826)
                (8, 0.22510964133590458)
                (9, 0.22558977851644155)
                (10, 0.22601561819761984)
                (11, 0.22633108137175445)
                (12, 0.2266590558439493)
                (13, 0.22696798956021666)
                (14, 0.2272422496937216)
                (15, 0.22749194601178155)
                (16, 0.2277364636547864)
                (17, 0.2279291799031198)
                (18, 0.22811005797982223)
                (19, 0.22828537061810497)
            };\label{plot_two}
            \end{axis}

            \begin{axis}[
                axis y line*=right,
                axis x line=none,
                ylabel=y-axis 2
                grid style={line width=.1pt, draw=gray!10},
                major grid style={line width=.2pt,draw=gray!50},
                minor tick num=2,
                height=2in,
                width=\linewidth,
                ylabel style= {align=center},
                ylabel={\color{mustard}{Human Eval}},
                legend style={cells={align=left}, at={(0.9,0.1)},anchor=south east},
            ]
            \addlegendimage{/pgfplots/refstyle=plot_two}\addlegendentry{PickScore}
            \addplot+[
                mark=square, line width=1mm, mark options={scale=3}, smooth,
                tension={0.5},
                mustard,
                error bars/.cd, y dir=both, y fixed=4,
                error bar style={line width=1mm,solid},
                error mark options={scale=3,line width=1mm,rotate=90}
            ] coordinates {
                (0,50)
                (9,81.0)
                (19,79.5)
            };
            \addlegendentry{Human Eval}
            \end{axis}
        \end{tikzpicture}%
        }
        \caption{
            {\bf Performance with number of optimization steps for captioning (left) and generation (right).}
            We show the
            \setulcolor{niceblue}
            \ul{optimization metric}, \ie \scorer{}'s output (CLIP Similarity and PickScore), and the
            \setulcolor{mustard}
            \ul{downstream metric} (SPICE~\cite{anderson2016spice} and human evaluation win\% for quality), respectively.
            Both tend to improve over optimization steps, and correlate with each other.
        }\label{fig:ablation:steps}
    \end{minipage}
    \hfill
    \begin{minipage}[t]{.29\textwidth}
        \resizebox{\linewidth}{!}{
        \begin{tikzpicture}
            \begin{axis}[
                grid=both,
                grid style={line width=.1pt, draw=gray!10},
                major grid style={line width=.2pt,draw=gray!50},
                minor tick num=2,
                axis x line*=bottom,
                axis y line*=left,
                height=1.3in,
                width=\linewidth,
                ylabel style= {align=center},
                ylabel={CLIP$^*$},
                xlabel={Initial set size},
                yticklabel style = {font=\small},
                xticklabel style = {font=\small},
                xmode=log,
                ytick={0.0,0.1,0.15},
            ]
            \addplot[mark=o, very thick, smooth, niceblue] plot coordinates {
                (30, 0.05405686956265708)
                (300, 0.1208553494438529)
                (3000, 0.15807164472341537)
                (30000, 0.17529554387181998)
            };
            \end{axis}
        \end{tikzpicture}%
        }
        \caption{
            {\bf Performance improves with size of initial set},
            suggesting it is crucial in bootstrapping the \generator. %
            CLIP$^*$: CLIP similarity at convergence.
        }\label{fig:ablation:init_set_size}
    \end{minipage}
\end{figure*}

\begin{table*}[t]
    \setlength{\tabcolsep}{2pt}
    \centering
    \resizebox{\linewidth}{!}{
    \begin{tabular}{|C{0.15in}|C{1.5in}|C{1.5in}|C{1.5in}|C{1.5in}|C{1.5in}|C{1.5in}|}
        \hline
        \# & 1 & 4 & 8 & 12 & 16 & 20 \\
        \hline
        \rotatebox[origin=r]{90}{Prompt rewrite}
        & {\it A red car and a white sheep.}
        & {\it A white sheep and a red car in a stunning, natural landscape with a few mountains and a lake.}
        & {\it A red car and a white sheep standing in a stunning, rural landscape with a red barn, a few trees, and a few birds.}
        & {\it A red car and a white sheep standing in a stunning, rural landscape with a red barn, a few trees, and a few birds.}
        & {\it A red car and a white sheep in a stunning, natural landscape with a few mountains, a few wildflowers, and a serene lake.}
        & {\it A red car and a white sheep in a stunning, natural landscape with a few mountains, a few wildflowers, and a serene lake.}
        \\
        \hline
        \rotatebox[origin=c]{90}{Generated Image}
        & \raisebox{-0.5\height}{\frame{\includegraphics[width=1.5in]{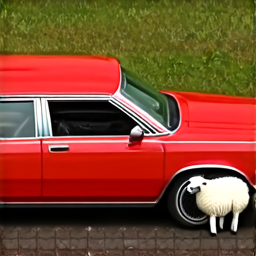}}}
        & \raisebox{-0.5\height}{\frame{\includegraphics[width=1.5in]{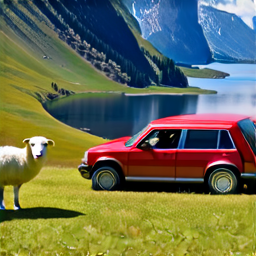}}}
        & \raisebox{-0.5\height}{\frame{\includegraphics[width=1.5in]{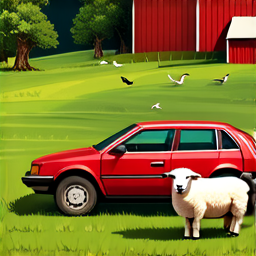}}}
        & \raisebox{-0.5\height}{\frame{\includegraphics[width=1.5in]{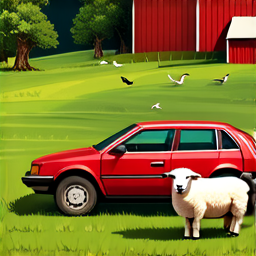}}}
        & \raisebox{-0.5\height}{\frame{\includegraphics[width=1.5in]{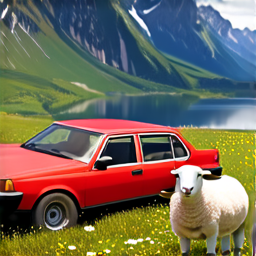}}}
        & \raisebox{-0.5\height}{\frame{\includegraphics[width=1.5in]{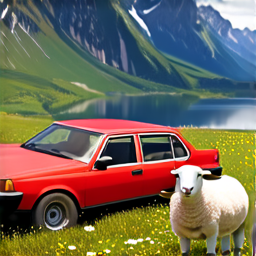}}}
        \\
        \hline
    \end{tabular}%
    }
    \captionof{figure}{
        \textbf{Improved image generation over optimization steps.}
        The quality of the output
        improves over the optimization steps (\#). We also show the prompt being
        produced by the LLM in the \generator, which is passed to the T2I model.
    }
    \label{fig:ablation:image_gen_by_steps}
\end{table*}

{\noindent \bf Results.}
\cref{fig:results:style_tx} shows some sample style transfer results.
\OURS generalizes to this novel task completely zero-shot and produces accurately stylized images.
Note that it achieves such edits not only without any training, but also without the LLM actually seeing any features from either the test sample, or the style image!

\subsection{Cross-Modal Arithmetic}
Finally, we explore an interesting application enabled by \OURS. %
Unlike prior work~\cite{kazemi2024we} that maps embeddings to continuous image space,
our gradient-free approach in \OURS enables inverting such embeddings into the
discrete text space instead.
This is also exemplified by our results in~\cref{sec:expts:image_caption,sec:expts:video_caption,sec:expts:audio_caption}.
This enables an interesting application of
cross-modal arithmetic.
We take inspiration from ImageBind~\cite{girdhar2023imagebind}, which mapped multiple different modalities into the image embedding space. Using this shared embedding, authors were able to combine modalities, and generate or retrieve images given that combination.
\OURS, in fact, is even more flexible, as inversion to text enables interfacing with many more models. For instance, ImageBind showed results on audio to image generation by leveraging a DALLE-2-like T2I model~\cite{ramesh2022hierarchical}. This was possible as ImageBind happened to be aligned to the CLIP embedding space, same as what was used in DALLE-2. As such, ImageBind was not compatible with any other T2I model, such as a latent diffusion model~\cite{rombach2022high}.
A textual representation, on the other hand, would work with any T2I model, including those that do not represent textual input as a point on an embedding space. In~\cref{fig:results:arithmetic}, we show examples of combining image and audio modalities. We first invert both image and audio into text using~\cref{sec:expts:image_caption} and~\cref{sec:expts:audio_caption} respectively, combine the two outputs using an LLM (details in the~\cref{sec:appdx:llm_prompts}),
and finally convert the prompt into a high quality image as described in~\cref{sec:expts:image_gen}. The resulting generated image combines the semantic concepts from both those modalities.

\subsection{Ablations}\label{sec:expts:ablations}

We now ablate some key design choices in \OURS. We primarily focus on the
image captioning task for most of the analysis, and improved image generation for some of the ablations.
For computational ease, we randomly sample 1000 images from MSCOCO for captioning, and
use the 200 prompt DrawBench set for image generation, as the test set for this analysis. We report all metrics including CLIP similarity and
PickScore averaged over these sets.

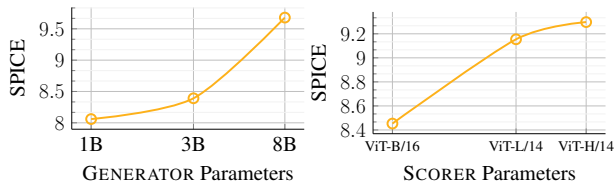
\begin{figure}[t]
    \resizebox{\linewidth}{!}{
    \begin{tikzpicture}
        \LARGE
        \begin{axis}[
            xtick={1, 3, 8, 70, 405},
            xticklabels={
                1B, 3B, 8B
            },
            grid=both,
            grid style={line width=.2pt, draw=gray!10},
            major grid style={line width=.4pt,draw=gray!50},
            minor tick num=2,
            axis x line*=bottom,
            axis y line*=left,
            height=2in,
            width=\linewidth,
            ylabel style= {align=center},
            ylabel={SPICE},
            xlabel={\generator Parameters}, %
            xmode=log,
        ]
        \addplot[mark=o, ultra thick, smooth, mustard, mark options={scale=2}] plot coordinates {
            (1, 8.0597)
            (3, 8.3913)
            (8, 9.6791)
        };
        \end{axis}
    \end{tikzpicture}%
    \begin{tikzpicture}
        \LARGE
        \begin{axis}[
            xtick={86, 307, 632},
            xticklabels={ViT-B/16, ViT-L/14, ViT-H/14},
            grid=both,
            grid style={line width=.1pt, draw=gray!10},
            major grid style={line width=.2pt,draw=gray!50},
            minor tick num=2,
            axis x line*=bottom,
            axis y line*=left,
            height=2in,
            width=\linewidth,
            ylabel style= {align=center},
            ylabel={SPICE},
            xlabel={\scorer Parameters}, %
            xticklabel style = {font=\large},
            xmode=log,
        ]
        \addplot[mark=o, ultra thick, smooth, mustard, mark options={scale=2}] plot coordinates {
            (86, 8.4532)
            (307, 9.1543)
            (632, 9.2974)
        };
        \end{axis}
    \end{tikzpicture}%
    }
    \caption{
        {\bf Impact of \generator (left) and \scorer (right) size.}
        We report the downstream metric SPICE for both these ablations since
        CLIP similarity may not be comparable across model sizes.
        For the \generator and \scorer we use different sized \llamaOnly{} or MetaCLIP models. As evident from the graphs, larger models perform better.
    }\label{fig:ablation:llm_size}
\end{figure}

{\noindent \bf Performance over optimization steps.}
We evaluate this for both the tasks in~\cref{fig:ablation:steps}. We report both
the \scorer output, which can be thought of as a ``training loss'' in our setup, as well as a downstream metric. We report SPICE~\cite{anderson2016spice} for image captioning, and human evaluation against the original prompt's generation for T2I. For the latter, we also show $\pm4$ point error bars, which we found to be the typical
random variance margins in human evaluations.
As~\cref{fig:ablation:steps} shows, both the \scorer outputs and the downstream metrics improve over optimization steps, converging after 10 to 20 steps.
We also note that the optimization objective (\scorer output)
correlates well with downstream performance.
Lastly, we visualize this result qualitatively in~\cref{fig:ablation:image_caps_by_steps,fig:ablation:image_gen_by_steps} for captioning and generation respectively. In both cases, the quality of the output improves over the steps, showing the effectiveness of \OURS.

{\noindent \bf Impact of the initial candidate set.}
We evaluate this in~\cref{fig:ablation:init_set_size}. We subsample the initial set to
different sizes, and see a strong positive correlation between that and the final performance.
This suggests that the initial bootstrapping set is critical to ensure the \generator
can produce sufficiently diverse candidates and avoid local minimas.

{\noindent \bf Size of the \generator and \scorer.}
We evaluate the effect of the size (in parameters) of the \generator (\llamaNumOnly{3}) and \scorer (MetaCLIP~\cite{xu2024demystifying}) in~\cref{fig:ablation:llm_size}, for image captioning.
We find larger models generally performed better, with LLM scaling exhibiting the most promising
gains. %
We also experiment with different kinds of \generator{}s and \scorer{}s in the~\cref{sec:appdx:addl_qual}.

\section{Conclusion and Future Work}

We have presented \OURS, a simple approach for solving multimodal tasks
without requiring any task specific
data curation or training.
\OURS exhitbits emergent zero-shot generalization to various different
tasks and modalities.
Notably, we show \OURS
obtains strong performance on captioning across three modalities: images, videos and
audio, showing that LLMs can see and hear without any training!
This further leads to improving and enabling various media generation tasks, such as image generation, image editing (style transfer) and cross-modal arithmetic.

While promising, \OURS has some limitations that future work can attempt to
address. Its performance is bounded by the ability of the \generator to
generate diverse candidates, and the \scorer to provide accurate feedback
to the \generator.
For instance, style transfer performance is limited by the resolution of Gram matrix distance in detecting %
fine-grained texture similarities, and the LLM's ability to describe potential styles.
As the LLMs and the multimodal models %
continue to improve~\cite{openai2024o1,fang2023data}, \OURS would improve with it.
Another limitation is the speed of the optimization process.
This would improve as the core LLMs become faster and more efficient,
and as their context length~\cite{munkhdalai2024leave} and reasoning abilities~\cite{openai2024o1} improve,
requiring fewer optimization steps.
It would also be interesting to apply
\OURS to other modalities and tasks, such
as for spatial and 3D tasks. %

\section*{Impact Statement}

Our work builds on top of LLMs and other multimodal models, and as such, the strengths and weaknesses of such models would be reflected when used with \OURS.
This includes the biases present in such models. However, \OURS being a training-free approach,
does not learn any new biases, and can easily be upgraded with better and less-biased LLMs
or multimodal models as they become available, without needing any re-training.
While we do not foresee any negative societal impact directly enabled by \OURS, any real-world deployment should consider additional safety evaluations before deployment.

\bibliography{main}
\bibliographystyle{icml2025}

\newpage
\appendix
\clearpage
\setcounter{page}{1}

\section{Human evaluation Setup}\label{sec:appdx:human_eval}

\begin{figure*}[t]
    \resizebox{\linewidth}{!}{
        \includegraphics[width=0.482\textwidth]{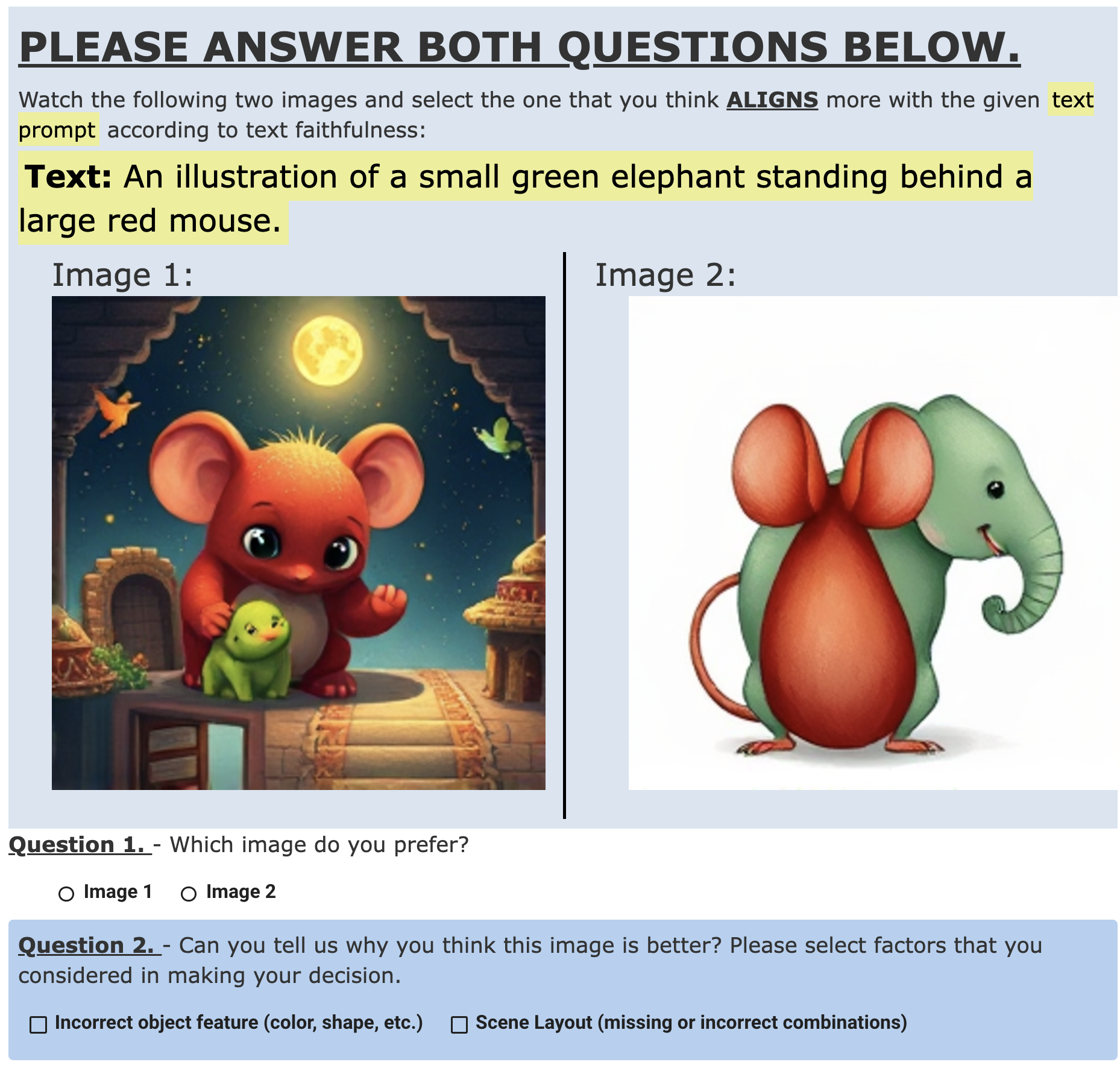}
        \includegraphics[width=0.518\textwidth]{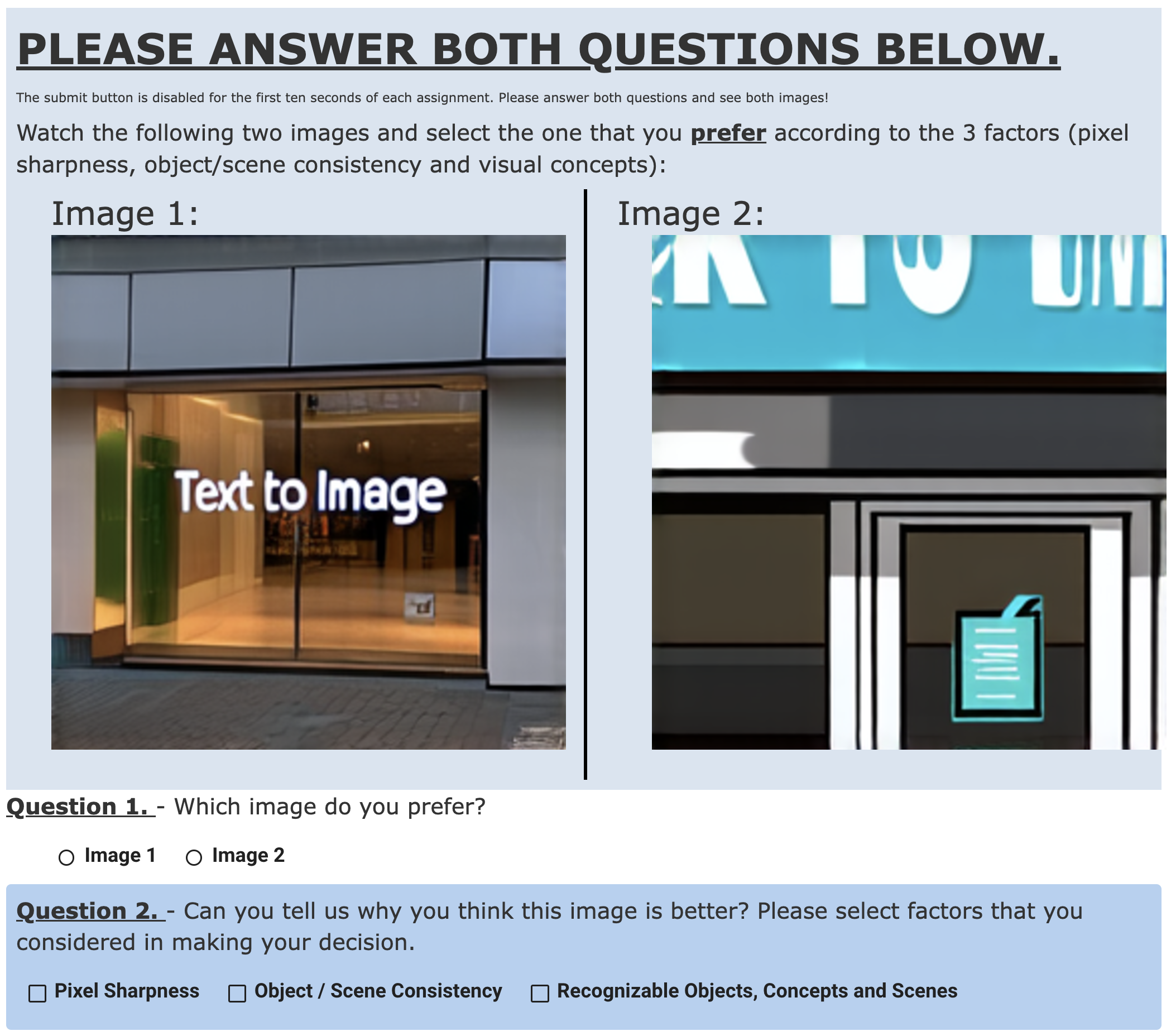}
        }
    \caption{
        {\bf Human evaluation question form for faithfulness (left) and quality (right).}
        We ask two questions. The first question asks which of the two images the rater prefers. The next question
         elaborates on the first response by asking the rater to list the reason for their preference, based on JUICE~\cite{girdhar2024emu} evaluation protocol.
    }\label{fig:supp:amt_interface}
\end{figure*}

\begin{figure}[t]
    \resizebox{\linewidth}{!}{
    \includegraphics[width=\textwidth]{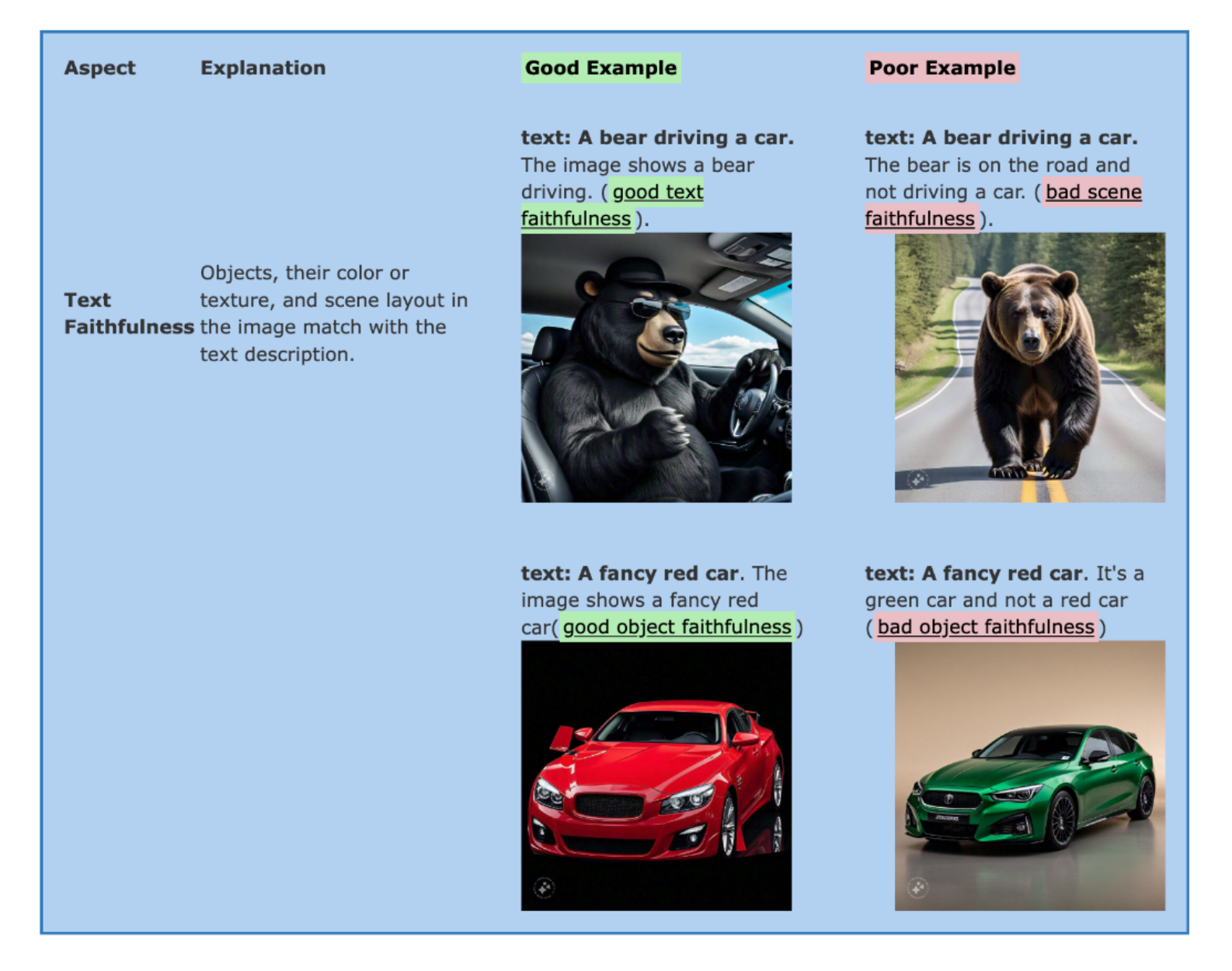}
    }
    \caption{
        {\bf High-quality image generation faithfulness examples.}
    We provide these examples to the annotators to judge the faithfulness. The generated image should have the correct object, color, texture or scene layout.
    }\label{fig:supp:examples_faithfulness}
\end{figure}

\begin{figure}[t]
    \resizebox{\linewidth}{!}{
    \includegraphics[width=\textwidth]{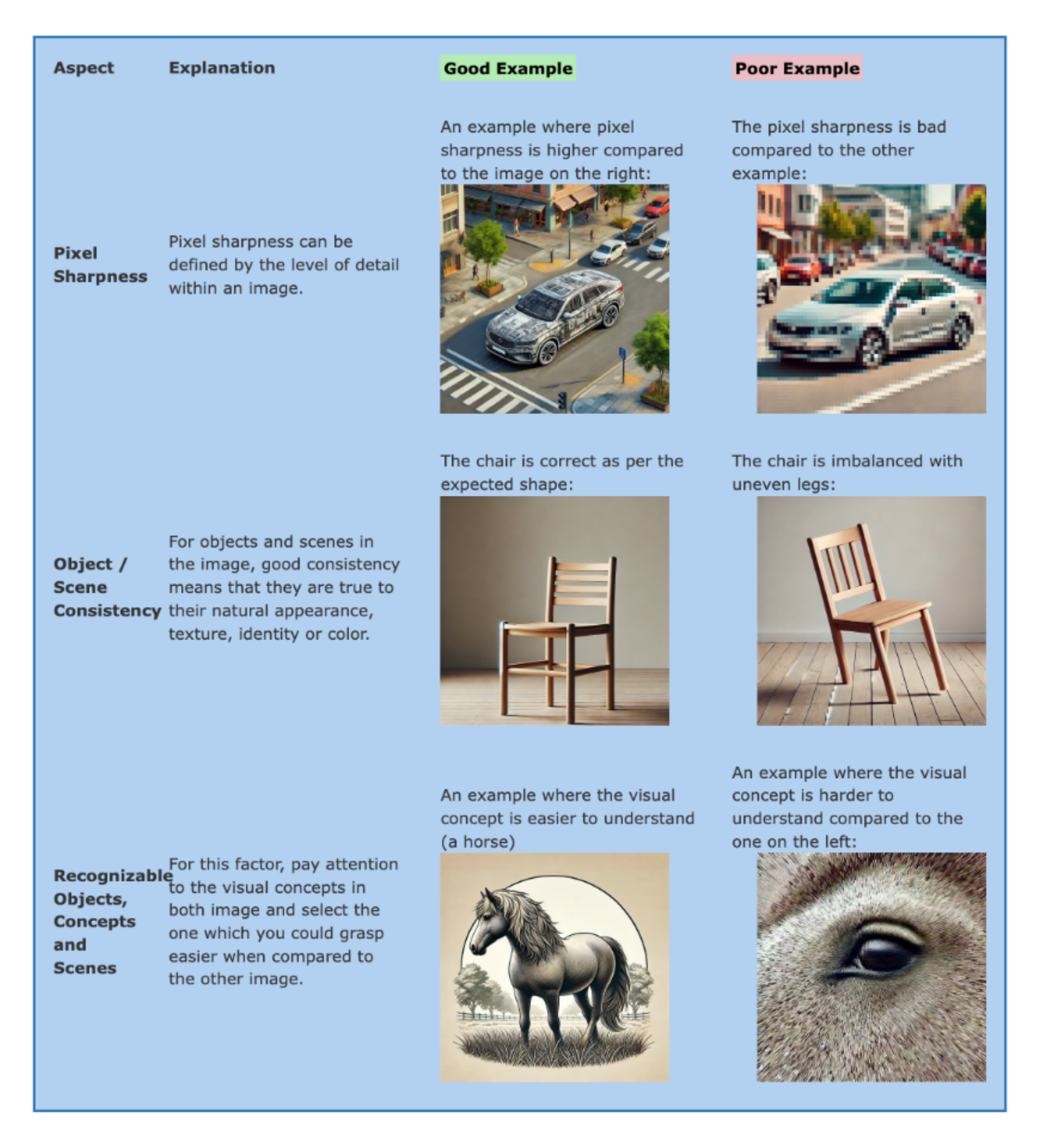}
    }
    \caption{
        {\bf High-quality image generation quality examples.}
    We provide these examples to the annotators to judge the quality. The resulting image should have high pixel sharpness, object and scene consistency, and the concepts should be recognizable.
    }\label{fig:supp:examples_quality}
\end{figure}

We use Amazon Mechanical Turk (AMT) for human evaluation, with a setup similar to~\cite{girdhar2024emu}. As discussed in~\cref{sec:expts:image_gen}, we evaluate the enhanced image generation on two axes: quality and faithfulness to the text prompt. Screenshots of the evaluation interface are shown in~\cref{fig:supp:amt_interface}.
The annotators are also given elaborate instructions with examples, as shown in~\cref{fig:supp:examples_faithfulness,fig:supp:examples_quality} for the faithfulness and quality evaluations respectively.
We give each sample to three annotators and use majority vote to choose the winning model.

\section{LLM prompts}\label{sec:appdx:llm_prompts}

\subsection{Generating the initial candidate set}

We bootstrap captioning tasks with an initial candidate set.
We use the following strategies to create this initial candidate set.

\noindent \textbf{Image and video captioning:} We use the initial candidate set as generated in~\cite{gandelsman2024interpreting}. See Table 7 in their Appendix for the exact prompt. The overall idea is to take class labels from ImageNet~\cite{deng2009imagenet} and prompt the LLM to generate 40 candidate captions based on the concept of the chosen class label. This process yields a set of around $30,000$ captions. This same set is used for the video captioning task.

\noindent \textbf{Audio captioning.} Since audio captions typically describe audible information, as opposed to visual information, we recreate the candidate set based on the above protocol with small modifications. We use class labels from AudioSet~\cite{gemmeke2017audio} and generate $50$ captions per audio label. We use Llama 3.1 %
and provide the following prompt:

\begin{tcolorbox}[breakable, boxrule=0.2mm]
    Generate 50 diverse descriptive captions for an audio clip that features the sound of \{class\_label\}. Write a concise and vivid description of what can be heard in the clip, using complete sentences. For example:

1. A car drives by with its horn honking.

2. Children are playing and laughing in a park.

3. Heavy rain falls on pavement and roofs.

4. A crowd cheers and applauds at a sports event.

Write the generation as if a person would write that after listening to the audio clip. Do not mention concepts that cannot be heard, like sunshine, star, any color or taste.
Try to capture the main sounds and any background or accompanying noises in your caption, without referencing the fact that you're listening to an audio clip. Simply describe what can be heard.
Put each description in a different line, with a counter at the beginning (e.g. `1. ...'), don’t explain why, and don’t combine two different concepts (with `or' or `and'), and keep it short 15-20 words.
\end{tcolorbox}

where \{class\_label\} is the AudioSet class name.
The above prompt is similar to the prompt used in image and video captioning, adapted to mention the need for focus on audio. This process yields a candidate set of about $50,000$ captions.

\subsection{\generator prompts}

We use the following prompts in the \generator. Here, \{descriptions\} denote the list of previous generations, along with the score for each of the generation from the \scorer. The scores are sorted, and we provide one score and text per row. In high-quality image generation, we also provide an additional \{init\_description\} which is the original text prompt. \{requested\_number\} is the number of new generations that we ask from the \generator.

\noindent \textbf{Image Captioning:}

\begin{tcolorbox}[breakable, boxrule=0.2mm]
    You need to provide a short image description.
    I am providing to you a list of short image descriptions and scores.
    Higher score means that the image description characterizes the image better:

    \{descriptions\}

    Generate additional \{requested\_number\} short image descriptions that you think that will maximize the score and fully capture the image.
    Be concrete and try to find elements that are unique to this image.
    You can introduce new elements to the descriptions, combine unique elements and objects from provided descriptions to form new descriptions, rephrase individual descriptions, drop elements, or simplify descriptions.
    Be creative and don't be afraid to come up with erroneous descriptions. Put each description in a different line, with a counter at the beginning (e.g. ``1. ..."), and try to keep them very short (up to 10 words).
\end{tcolorbox}

\noindent \textbf{Video captioning:}
\begin{tcolorbox}[breakable, boxrule=0.2mm]
    You need to provide a short video description.
    I am providing to you a list of short video descriptions and scores.
    Higher score means that the video description characterizes the video better:

    \{descriptions\}

    Generate additional \{requested\_number\} short video descriptions that you think that will maximize the score and fully capture the video.
    Be concrete and try to find elements that are unique to this video.
    You can introduce new elements to the descriptions, combine unique elements and objects from provided descriptions to form new descriptions, rephrase individual descriptions, drop elements, or simplify descriptions.
    Be creative and don't be afraid to come up with erroneous descriptions. Put each description in a different line, with a counter at the beginning (e.g.``1. ..."), and try to keep them short up to 20-25 words.
\end{tcolorbox}

\noindent \textbf{Audio captioning:}
\begin{tcolorbox}[breakable, boxrule=0.2mm]
    You need to provide a short audio description.
    I am providing to you a list of audio descriptions and scores.
    Higher score means that the audio description characterizes the audio better:

    \{descriptions\}

    Generate additional \{requested\_number\} short audio descriptions that you think that will maximize the score and fully capture the audio.
    Be concrete and try to find elements that are unique to this audio.
    You can introduce new elements to the descriptions, combine unique elements and objects from provided descriptions to form new descriptions, rephrase individual descriptions, drop elements, or simplify descriptions.
    Be creative and don't be afraid to come up with erroneous descriptions. Put each description in a different line, with a counter at the beginning (e.g. ``1. ..."), and try to keep them short (under 20 words).
\end{tcolorbox}

\noindent \textbf{High-quality image generation:}
\begin{tcolorbox}[breakable, boxrule=0.2mm]
    You need to expand and rephrase the provided description for image generation to make the best image, by maximizing the image score:
    The description is: \{init\_description\}

    Here are some example rephrases and the corresponding image scores:

    \{descriptions\}

    Generate additional \{requested\_number\} descriptions that will maximize the score. Be concrete and come up with different descriptions with various guesses for the possible way to rephrase and expand it, in a way that will maximize the score.
    You can introduce new elements to the descriptions, combine unique elements and phrasings from provided descriptions to form new ones, drop description parts, or simplify them.
    Be creative and don't be afraid to come up with erroneous descriptions. Put each instruction in a different line, with a counter at the beginning (e.g. ``1. ..."), and keep them short (less than 77 words).
\end{tcolorbox}

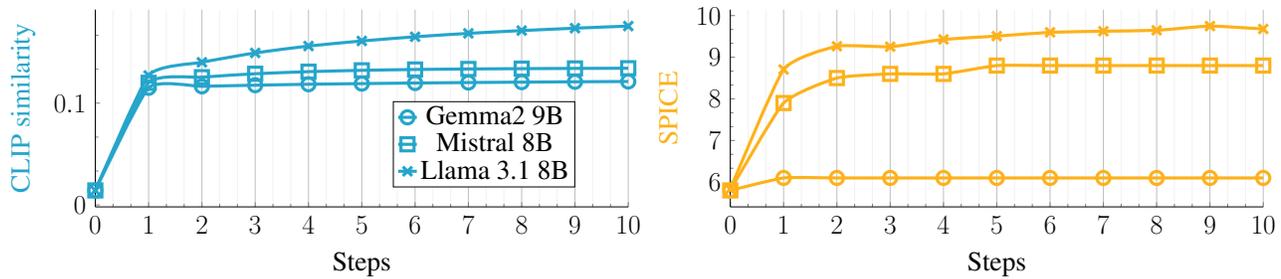
\begin{figure*}
    \begin{minipage}[t]{\textwidth}
        \Huge
        \resizebox{\linewidth}{!}{
        \hfill
        \begin{tikzpicture}
            \pgfplotsset{
                scale only axis,
                xtick distance=1,
                xmin=0, xmax=10
            }
            \begin{axis}[
                xmajorgrids,
                xminorgrids,
                grid style={line width=.1pt, draw=gray!10},
                major grid style={line width=.2pt,draw=gray!50},
                minor tick num=2,
                axis x line*=bottom,
                axis y line*=left,
                height=2.5in,
                width=\linewidth,
                ylabel style= {align=center},
                ylabel={\color{niceblue}{CLIP similarity}},
                xlabel={Steps},
                ytick={0.0,0.1,0.2},
                legend style={cells={align=left}, at={(0.9,0.1)},anchor=south east},
            ]
            \addplot[mark=o, line width=1mm, mark options={scale=3}, smooth, tension={0.3}, niceblue] plot coordinates {
                (0, 0.014238783842376881)
                (1, 0.11508447789217359)
                (2, 0.11654922129520819)
                (3, 0.11765777442473263)
                (4, 0.11857901867011085)
                (5, 0.11922192415550127)
                (6, 0.11980551315012888)
                (7, 0.12027135097501265)
                (8, 0.1206522461382588)
                (9, 0.12101566205971205)
                (10, 0.12129560552500002)
            };\label{plot_one}
            \addplot[mark=square, line width=1mm, mark options={scale=3}, smooth, tension={0.3}, niceblue] plot coordinates {
                (0, 0.01421688077264298)
                (1, 0.12002419658750296)
                (2, 0.12552588580921292)
                (3, 0.12892052876949311)
                (4, 0.13096412984281777)
                (5, 0.13227721277624369)
                (6, 0.13310740973055363)
                (7, 0.13360806154459715)
                (8, 0.13392560188472272)
                (9, 0.13417325086891652)
                (10, 0.13432766108959912)
            };\label{plot_two}
            \addplot[mark=x, line width=1mm, mark options={scale=3}, smooth, tension={0.3}, niceblue] plot coordinates {
                (0, 0.01421687289538022)
                (1, 0.1272442454956472)
                (2, 0.1401589315906167)
                (3, 0.14938828766345977)
                (4, 0.1561347666159272)
                (5, 0.16111914109438658)
                (6, 0.16525025141984223)
                (7, 0.16861245894432067)
                (8, 0.1713299216926098)
                (9, 0.17369869603961707)
                (10, 0.17575965739041566)
            };\label{plot_three}
            \addlegendimage{/pgfplots/refstyle=plot_one}\addlegendentry{Gemma2 9B}
            \addlegendimage{/pgfplots/refstyle=plot_two}\addlegendentry{Mistral 8B}
            \addlegendimage{/pgfplots/refstyle=plot_three}\addlegendentry{Llama 3.1 8B}
            \end{axis}

        \end{tikzpicture}%
        \hfill
        \begin{tikzpicture}
            \pgfplotsset{
                scale only axis,
                xtick distance=1,
                xmin=0, xmax=10
            }
            \begin{axis}[
                xmajorgrids,
                xminorgrids,
                grid style={line width=.1pt, draw=gray!10},
                major grid style={line width=.2pt,draw=gray!50},
                minor tick num=2,
                axis x line*=bottom,
                axis y line*=left,
                height=2.5in,
                width=\linewidth,
                ylabel style= {align=center},
                ylabel={\color{mustard}{SPICE}},
                xlabel={Steps},
            ]
            \addplot[mark=o, line width=1mm, mark options={scale=3}, smooth, tension={0.3}, mustard] plot coordinates {
                (0, 5.8)
                (1, 6.1)
                (2, 6.1)
                (3, 6.1)
                (4, 6.1)
                (5, 6.1)
                (6, 6.1)
                (7, 6.1)
                (8, 6.1)
                (9, 6.1)
                (10, 6.1)
            };\label{plot_one}
            \addplot[mark=square, line width=1mm, mark options={scale=3}, smooth, tension={0.3}, mustard] plot coordinates {
                (0, 5.8)
                (1, 7.9)
                (2, 8.5)
                (3, 8.6)
                (4, 8.6)
                (5, 8.8)
                (6, 8.8)
                (7, 8.8)
                (8, 8.8)
                (9, 8.8)
                (10, 8.8)
            };\label{plot_two}
            \addplot[mark=x, line width=1mm, mark options={scale=3}, smooth, tension={0.3}, mustard] plot coordinates {
                (0, 5.8314)
                (1, 8.7049)
                (2, 9.2636)
                (3, 9.2522)
                (4, 9.4283)
                (5, 9.5071)
                (6, 9.5984)
                (7, 9.6251)
                (8, 9.6475)
                (9, 9.7484)
                (10, 9.6791)
            };\label{plot_three}
            \end{axis}

        \end{tikzpicture}%

        }
        \caption{
            {\bf Performance comparison with various LLMs as \generator -- Gemma2 9B, Mistral 8B, and Llama 3.1 8B, on CLIP similarity (left) and SPICE (right).}
            This trend shows that the performance improves over optimization steps, regardless of the choice of the LLM. In particular, Llama 3.1 8B gives the best performance, and is used primarily as the \generator in our experiments.
        }\label{fig:supp:choice-of-llm-with-steps}
    \end{minipage}
\end{figure*}

\begin{table}[t]
    \centering
        \begin{tabu}{l|cc}
            \toprule[1pt]
            Model & Training data & SPICE  \\
            \midrule
            CLIP & WIT~\cite{radford2021learning} & 8.5  \\
            SigLIP & WebLI~\cite{zhai2023sigmoid} & \textbf{9.7}  \\
            MetaCLIP & MetaCLIP 400M~\cite{xu2024demystifying} & 9.2  \\
            DFN & DFN2B~\cite{fang2023data} & 9.3  \\
            \bottomrule[1pt]
        \end{tabu}%
    \caption{{\bf Type of \scorer.} All the models are based on ViT-L/14 with different training data.
    }
    \label{table:ablation:clip_type}
\end{table}

\noindent \textbf{Style transfer:}
\begin{tcolorbox}[breakable, boxrule=0.2mm]
    You need to generate instructions for image editing that minimize a pair of scores:
    I am providing you a list of example editing instructions and their pairs of scores.

    \{descriptions\}

Generate additional \{requested\_number\} editing instructions. Be concrete and come up with different instructions with various guesses for the possible edits that will minimize both of the scores.
You can introduce new styles to the instructions, combine unique styles and textures from provided instructions to form new instructions, rephrase individual instructions, drop instruction parts, or simplify them.
Be creative and don't be afraid to come up with erroneous editing instructions. Put each instruction in a different line, with a counter at the beginning (e.g. ``1. ..."), and keep them short (less than 50 words).

\end{tcolorbox}

\noindent \textbf{Cross-modal arithmetic:}
\begin{tcolorbox}[breakable, boxrule=0.2mm]%
I have an image description and an audio description that I want to combine together into a text description that will help an AI imagine that scene. As an example, if the caption says ``Crane on a grass" and the audio says ``An ocean with the waves crashing on shore" then you need to generate a text description like ``Crane beside the shore with waves coming". The combinations can be imaginative and not necessarily true in real world.
Here are the captions and the audio description: \\
Image caption: \{image\_caption\}\\
Audio caption: \{audio\_caption\}\\
Generate the combined caption in a single sentence.
\end{tcolorbox}

\section{Additional Results and Ablations}\label{sec:appdx:addl_qual}
\subsection{Video Captioning}
Please see the attached supplementary video for video captioning results.

\subsection{Audio Captioning}
Please see the attached supplementary video for audio captioning results.

\subsection{Cross-Modal Arithmetic}
Please see the attached supplementary video for cross-modal arithmetic results.

\subsection{Ablations}

\paragraph{Different types of \generator and \scorer}
We experiment with different open source LLMs, and CLIP models.
We present our results in~\cref{fig:supp:choice-of-llm-with-steps,table:ablation:clip_type}.

First, we observe
that while we see the same trend of performance improvement over steps for all LLMs we considered, the Llama model performed the best, with Mistral a close second. This shows that \OURS generalizes to various different LLMs, and can expect to improve as LLMs advance further.

Second, we see in~\cref{table:ablation:clip_type} that SigLIP is the strongest scorer for this task, when compared with models of the same parameter size. Again, the performance remains high for all scorers, showcasing the flexibility of \OURS.

\end{document}